%% file: main.tex
\documentclass[10pt,twocolumn,letterpaper]{article}

 \usepackage{cvpr}              
\usepackage{graphicx}
\usepackage{amsmath}
\usepackage{amssymb}
\usepackage{booktabs}
\usepackage{multirow}
\usepackage{mathtools}
\usepackage{float}
\usepackage{todonotes}
\usepackage{tikz}
\usepackage{algorithm}
\usepackage{algpseudocode}


\definecolor{cvprblue}{rgb}{0.21,0.49,0.74}
\usepackage[pagebackref,breaklinks,colorlinks,allcolors=cvprblue]{hyperref}

\newcommand{\PAR}[1]{\noindent {\bf #1~}}

\def\paperID{11976} 
\def\confName{CVPR}
\def\confYear{2025}

\title{COBRA - COnfidence score Based on shape Regression Analysis for quality assessment of object pose estimation}

\author{Panagiotis Sapoutzoglou$^{1,2}$ \;\; Georgios Giapitzakis$^1$ \;\; Georgios Floros$^3$ \;\; George Terzakis$^2$ \;\; \\ Maria Pateraki$^{1,2}$ \\
\small\normalfont $^1$National Technical University of Athens \quad $^2$ Institute of Communication \& Computer Systems \\ 
\small\normalfont \quad $^3$ Independent researcher}

\begin{document}
\maketitle
\input{sec/0_abstract}    
\input{sec/1_intro}

\input{sec/2_Related_work}
\input{sec/3_Method}
\input{sec/4_Experiments}
\input{sec/5_Conclusion_Future_Work}
{
    \small
    \bibliographystyle{ieeenat_fullname}
    \bibliography{main}
}
\clearpage
\appendix
\input{sec/Supplementary} 


\end{document}

%% file: sec/0_abstract.tex
\begin{abstract}
We propose a generic procedure for assessing 6D object pose estimates. Our approach relies on the evaluation of discrepancies in the geometry of the observed object, in particular its respective estimated back-projection in 3D, against a putative functional shape representation comprising mixtures of Gaussian Processes, that act as a template. Each Gaussian Process is trained to yield a fragment of the object's surface in a radial fashion with respect to designated reference points. We further define a pose confidence measure as the average probability of pixel back-projections in the Gaussian mixture. The goal of our experiments is two-fold. a) We demonstrate that our functional representation is sufficiently accurate as a shape template on which the probability of back-projected object points can be evaluated, and, b) we show that the resulting confidence scores based on these probabilities are indeed a consistent quality measure of pose.
\end{abstract}

%% file: sec/1_intro.tex
\section{Introduction}
\label{sec:intro}

\begin{figure}[htbp]
    \centering
    \includegraphics[width=0.9\columnwidth]{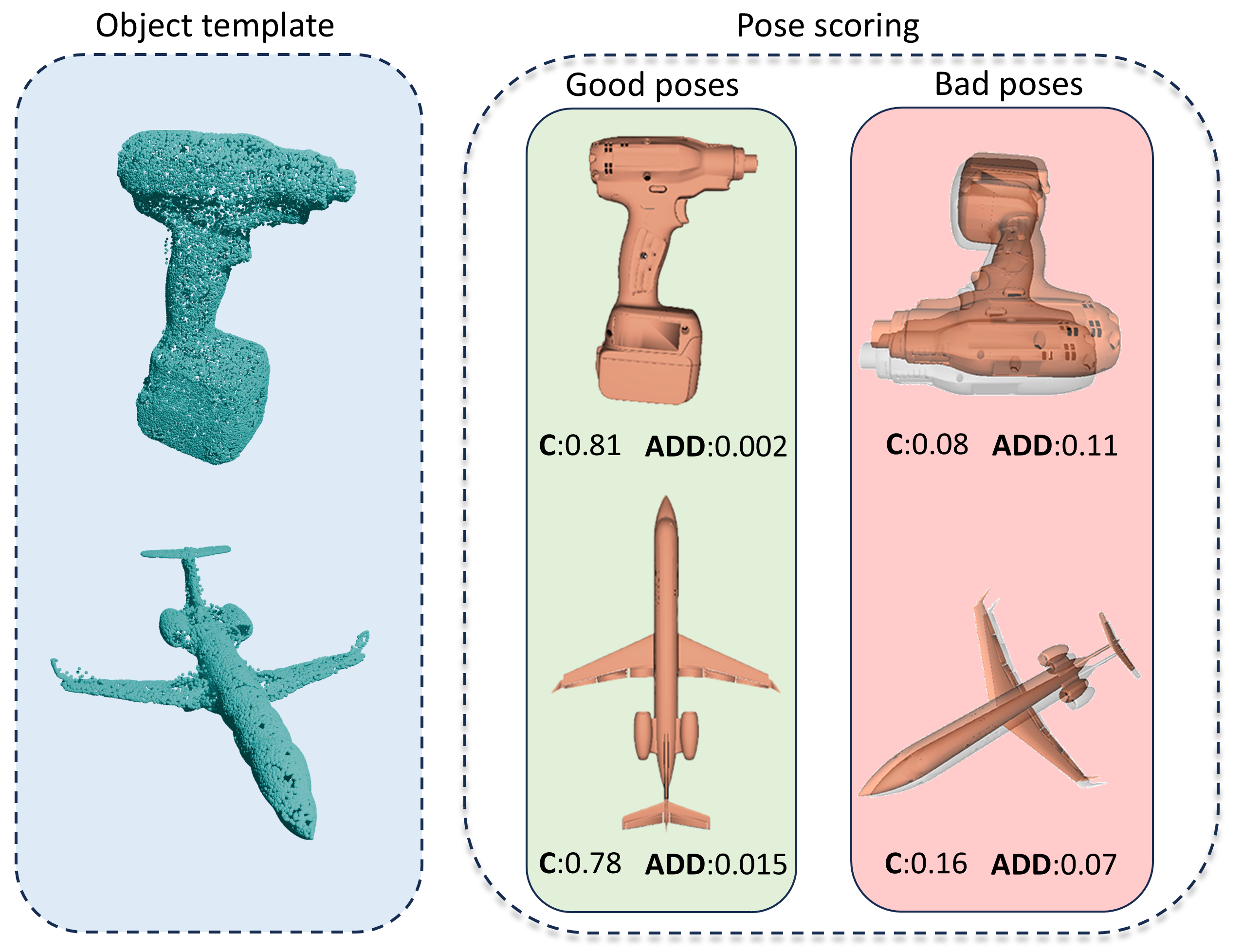}
    \caption{Our lightweight, GP-based representation (left) has the capacity to capture the shape variability of complex real-world objects, thus providing a reliable confidence score for 6D pose estimates (right).}
    \label{fig:main}
\end{figure}

The problem of recovering the pose of an object arises in a variety of applications in robotics and computer vision and constitutes a topic of ongoing scrutiny. In the case of rigid objects, the problem is reduced to the estimation of the object's 3D location and 3D orientation in the coordinate system of the camera, also referred to as 6D object pose. With recent developments in machine learning, neural network based solutions~\cite{sundermeyer,2021Asurvey,Zhu2022} have dominated the field by demonstrating significant accuracy, particularly in terms of estimating the 6D pose of objects from a single image. By employing latent representations, these methods have evolved to utilize different types of input data alongside images to address more challenging scenarios, such as multiple objects in the scene~\cite{hodan2020epos}, weak textures~\cite{jin2022PlosOne}, symmetries~\cite{cai20223DV} and unseen objects~\cite{nguyen2023nope}, in a fully supervised~\cite{lin2021} or a semi-supervised manner~\cite{Lin2022}.

This paper aims to address a key challenge in object pose estimation, evaluating the quality of pose predictions in a consistent and method-agnostic way. Our motivation stems from the need for a reliable method to evaluate the confidence of 6D pose estimates, particularly in applications involving robotic grasping and/or 3D object tracking, where precise and reliable pose information is critical for successful interactions with objects. While methods such as~\cite{nguyen2023nope,hodan2020epos,rambach2018learning} perform well in terms of estimating object poses in challenging scenarios, they often lack a robust way of measuring the uncertainty of predictions at runtime. Most existing confidence measures are intrinsic to the prediction model and are not capable to generalize across different methods. To address this issue, \cite{tremblay18} emphasizes the importance of uncertainty quantification, underlining the necessity of integrating confidence measures to handle real-world variability and complexity. Moreover, \cite{hodan2020epos} highlights the challenges in self-assessed confidence measures, pointing out the need for method-independent metrics.

\begin{figure*}[t]
    \centering
    \includegraphics[width=\textwidth]{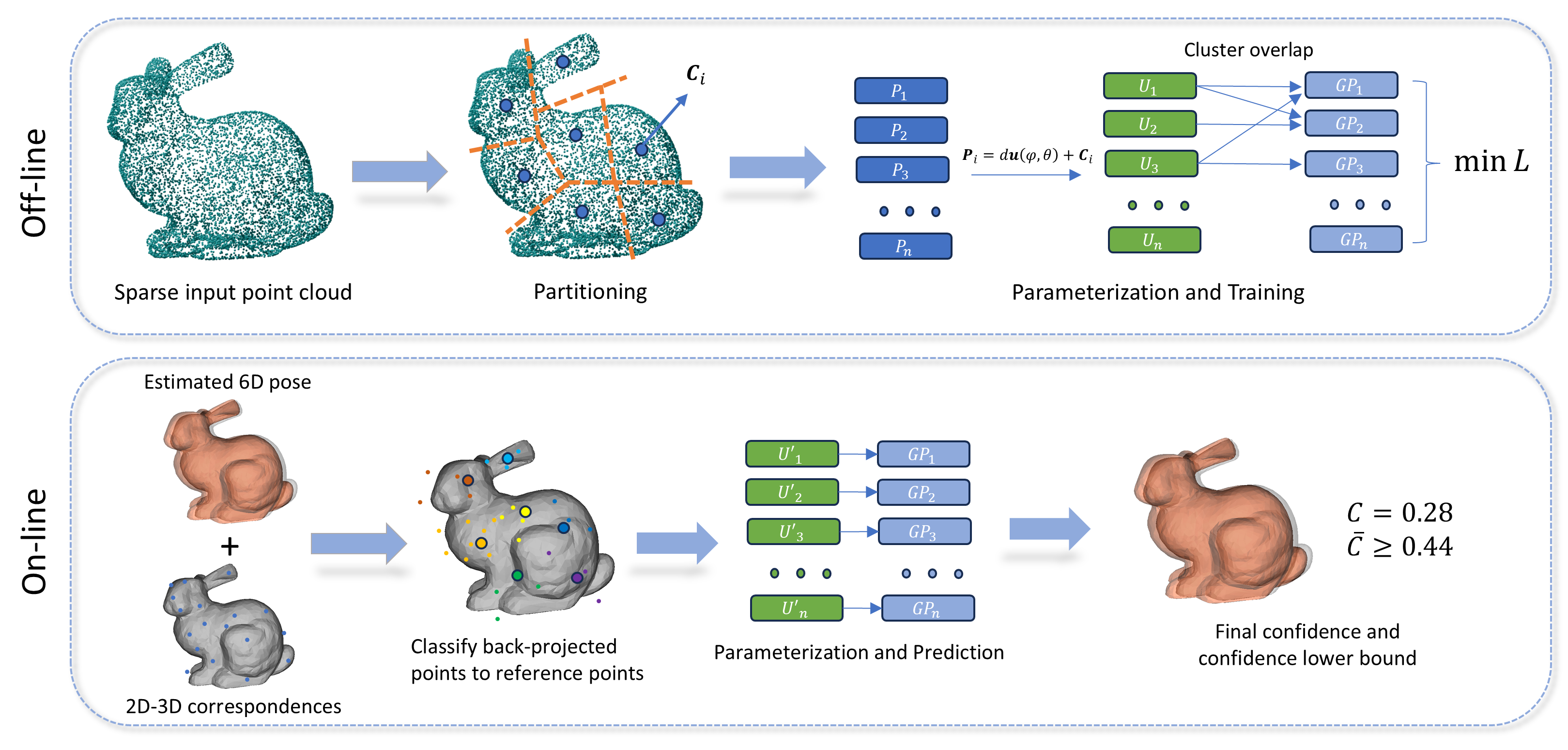}
\caption{\textbf{Off-line stage}: We partition a sparse point cloud using distance-based clustering and extract reference points (Sec.~\ref{ssect:lightweighht_shape_templates}). Each 3D point $P_i$ is parameterized as a bearing vector in spherical coordinates $U_i$ relative to its reference point $C_i$. The directions, with distances as targets, serve as inputs to a Gaussian Process ($GP_i$) (Sec.~\ref{ssect:GP_prior_for_distance}). The GP mixture model forms our shape representation (template). \textbf{On-line stage}: Using an estimated 6D pose and 2D-3D correspondences, we back-project the object's pixels, parameterize them, and predict distances using the template. Finally, a confidence score (Eq.~\ref{eq:generic_confidence}) and a lower bound (Eq.~\ref{eq:confidence_bound}) are computed to assess the pose quality.}
\end{figure*}

In our solution, we employ a lightweight shape representation based on Gaussian Process (GP)~\cite{Rasmussen2004, hida1993gaussian,williams2006gaussian} mixture models. Using regression model mixtures is a commonly employed strategy, take for instance~\cite{has2023gradient}. Our approach enables the estimation of a confidence score by evaluating the geometric consistency between putative shape representations of objects (henceforth referred to as ``templates'') and reconstructions obtained from 3D-2D correspondences. To construct the template each GP is trained on a specific region of the target surface. We draw inspiration from~\cite{dragiev2011}, that uses GPs to generate a probabilistic shape estimation integrating  visual, laser, and tactile data in the context of robotic grasping, and repurpose the model by training it to learn a directional distance field (DDF) over a spherical domain. Unlike methods that utilize DDFs along with a neural architecture~\cite{Tretschk2020, Armstrong2022, feng2022, Yenamandra2024}, we abstain from using any neural component as we require the template to be directly interpretable in geometric terms, for it is subsequently used as the foundation of our pose scoring module.

We introduce COnfidence score Based on shape Regression Analysis (COBRA), a consistent, method-agnostic confidence measure for pose estimates relying on shape discrepancy in the local object coordinate frame. The contribution of our method is three-fold.
\begin{enumerate}
    \item We propose a consistent and unbiased measure of pose estimates' quality by assessing the consistency of the object's back-projected image against a sufficiently accurate functional shape template.
    \item We present a novel, functional shape representation as a mixture of GP based priors. This representation is ``lightweight'' in the sense that it is derived from sparsely sampled points on the surface of the object. The accuracy of the representation is sufficient to yield reliable confidence values. 
    \item We evaluate the representation capabilities of the proposed model using both synthetic and real data and compare it against state-of-the-art competitors. Furthermore, we demonstrate the reliability of our scoring pose method through insightful ablation studies.
\end{enumerate}

%% file: sec/2_Related_work.tex
\section{Related work}

As our proposed method addresses two distinct problems, shape representation and quality evaluation of pose estimates, we discuss relevant literature for each separately.

\subsection{Shape representation}
\label{sec:sota_shape}
Shape representation constitutes a fundamental problem in 3D computer vision and has been approached by a variety of methodologies~\cite{cremers2015image}. Point-clouds, meshes, and voxels are the most popular \textit{explicit} representations able to capture arbitrary amounts of detail in the objects' shapes. They are used either as a standalone model or in conjunction with an encoder-decoder architecture~\cite{achlioptas2017learning, Wang2018_pixel2mesh, Brock2016}. Typical shortcomings of these representations include increased storage requirements for higher detail levels, and the need for special lookup structures to support operations such as nearest neighbor queries. 

On the other hand, \textit{implicit} representations are typically parametric functional models that are more storage-efficient and have proven to be adept at learning complex topologies~\cite{Farshian2023}. A line of research attempts to implicitly represent the object's surface, either as a continuous volumetric field~\cite{park2019deepsdf} or as the decision boundary of a classifier~\cite{Mescheder2019}. Other approaches use neural rendering to learn the object representation, optimizing a neural network~\cite{Mindenhall_NeRF}, a sparse grid of spherical harmonics~\cite{fridovich2022plenoxels} or an unstructured set of 3D Gaussians~\cite{kerbl20233d}. Local kernel functions~\cite{williams2021nkf, huang2023nksr} have also been used as a building block for implicit shape representations, focusing on the problem of surface reconstruction from sparse point clouds. Several extensions have been proposed, including the use of hierarchical structures~\cite{Paschalidou2020LearningUH} and the deformation of the underlying model~\cite{zheng2021deep}.

Overall, neural shape representations bear significant benefits, such as generalization, continuity in the input domain and storage efficiency. However, they are not purposed to reproduce details of specific shapes but rather as more general prediction frameworks. As such, they require swaths of data for training, rendering them unappealing for generating templates. 

Although Gaussian processes are prominent tools for non-parametric regression~\cite{titsias2009, williams2006gaussian}, to the best of our knowledge, only the work of Dragiev et al. \cite{dragiev2011} has utilized them specifically to refine shape uncertainty by integrating sensor data incrementally. Regardless, they have been extensively used for pattern learning from sparse data and recently, neural GP ensembles~\cite{damianou2013deep} or GP inspired architectures~\cite{garnelo2018, wilson2016} have been proposed. The latter boast superior performance on tasks such as orientation estimation from face images or digit magnitude prediction.

\subsection{6D pose estimation metrics and uncertainty}
\label{sec:sota_metrics}
The most prominent approaches adopted in literature for evaluating 6D pose estimation methods rely on available ground truth data. State of the art metrics such as Visible Surface Discrepancy (VSD), Maximum Symmetry-Aware Surface Distance (MSSD), Average Distance for distinguishable (ADD)~\cite{hoda} or 3-D IoU~\cite{wang2019} belong in this category. When 6D pose estimation is exploited in perception-to-action robotic tasks such as grasping, the rates of successful grasps are commonly computed as post-action performance indicators. 
However, in order to proactively avoid executing a potentially unnecessary grasping plan that was based on a weak pose estimate, a confidence measure over the object pose estimate is required. 
Works pertaining to the evaluation of the quality of pose estimates fall under the broader research area of \emph{uncertainty quantification} in robotics, or more particularly, \emph{predictive uncertainty}, wherein a measure of ``action risk'' is computed in order to reflect uncertainty in the underlying pose estimate. The majority of works that provide such measures incorporate self-assessment abilities with respect to the quantification of the uncertainty of their own predictions by directly training the model to output a confidence score~\cite{tremblay18,loquercio2020}. In principle, these are data-dependent approaches based on learning, and will naturally suffer from inherent bias that can be traced back to the training data.  
In other words, they exhibit limited robustness~\cite{blum_fishyscapes_2021} and tend to yield overconfident estimates~\cite{Shi2021} for which, a deviation from the ground-truth uncertainty is often clearly observed.
The method of~\cite{Shi2021} exploits ensemble methods~\cite{Lakshminarayanan2017}, which tend to be more lightweight in terms of inference time. In~\cite{Shi2021}, the problem is approached by an ensemble of different models optimizing translation/rotation/ADD errors and computing the average residuals as the uncertainty measure, though deviation from the true uncertainty remains.


%% file: sec/3_Method.tex
\section{Method}
\label{sect:method}

Our approach leverages 3D object templates to assess the fidelity of 2D-3D correspondences and poses provided by a given pertinent method. By comparing these templates to 3D reconstructions from a query image, we obtain an evaluation metric that is insensitive to the inference process.  

Methods able to detect objects in single images and predict their pose typically ship with measures of self-assessed confidence that provide some indication of quality for the estimated pose. However, these quality measures bear the inherent bias of the self-assessment process and can therefore be misleading for the back-end application. To circumvent this bias, we resort to the use of lightweight object templates to validate the geometric consistency of the estimated object pose. Put differently, we utilize the pose estimate to generate a partial 3D reconstruction of the detected object and establish a confidence score by assessing the deviations of the reconstructed points from the template surface.

\subsection{Shape templates with Gaussian Processes}
\label{ssect:lightweighht_shape_templates}

\begin{figure}[t]
     \centering
     \includegraphics[scale=0.15]{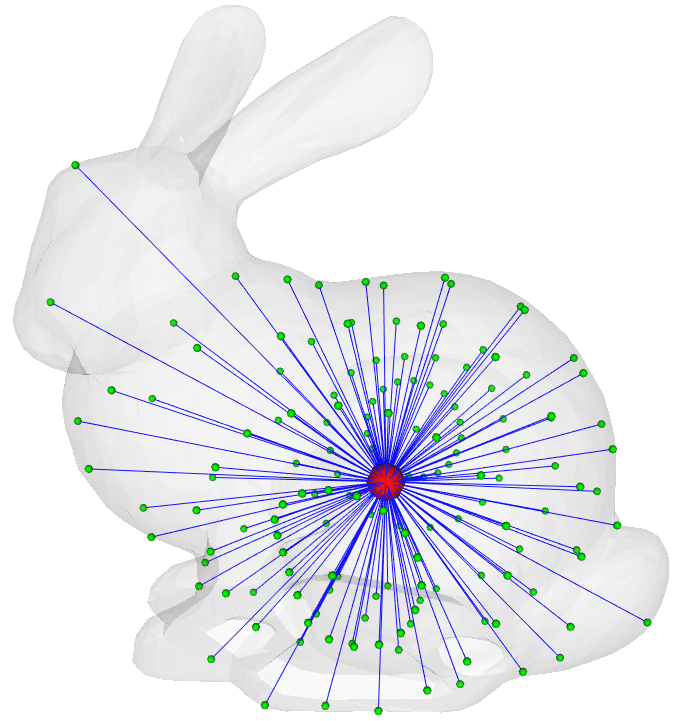}
     \caption{Spherical directional distance field (blue rays) centered at reference point $\pmb{C}$ (red).} 
     \label{fig:bunny_rays}
     \vspace{-1.5em}
\end{figure}

We construct our shape templates by learning directional distance fields on spherical coordinates domains affixed to suitable reference points in the object's local coordinate frame. Consider a reference point $\pmb{C}$ in the object's local frame and a given point $\pmb{P}$ on the object's surface. We parametrize $\pmb{r}=\pmb{P}-\pmb{C}$ in terms of its length and bearing (see Fig.~\ref{fig:bunny_rays}) as follows:
\begin{equation}
    \pmb{r}=\pmb{P}-\pmb{C} \coloneq \left(\phi, \theta,d \right),
    \label{eq:parametrization}
\end{equation}
where $\phi, \theta$ are the spherical coordinates of the bearing vector  $\pmb{r}/\Vert\pmb{r}\Vert$ and $d=\Vert \pmb{r}\Vert$ is the Euclidean norm of $\pmb{r}$. Thus, if $\pmb{u}(\phi, \theta)  \in \mathbb{R}^3$ is the bearing vector defined by the spherical coordinates $\phi$ and $\theta$, we may obtain $\pmb{P}$ as,
\begin{equation}
    \pmb{P}=\pmb{r}+\pmb{C}=d\,\pmb{u}(\phi, \theta) + \pmb{C}.
    \label{eq:point_from_parameters}
\end{equation}
The distance-direction parametrization of Eqs.~\eqref{eq:parametrization}, \eqref{eq:point_from_parameters}, bears a two-fold benefit. For a given reference point, $\pmb{C}$, in the interior of an object represented by a closed surface, we may represent all or part of the surface as a collection of distances along every possible direction from $\pmb{C}$.

\subsubsection{Gaussian Processes for distance prediction}
\label{ssect:GP_prior_for_distance}
Consider an object with an exterior surface that comprises points reachable by means of ray-casting from a point, $\pmb{C}$, in the interior of the object. To fit this surface to a function, we model distances from $\pmb{C}$ to the surface along a given direction $\pmb{u}(\phi, \theta)$, as a Gaussian process~\cite{Rasmussen2004, bishop2006pattern, perez2013gaussian},
\begin{equation}
    d \sim \mathcal{GP}\left(0, k\right), \;\;\text{s.t.}\;\; k : \left([0, \pi]\times[0, 2\pi]\right)^2\rightarrow \mathbb{R},
    \label{eq:gaussian_process}
\end{equation}
where $k$ is the so-called covariance function mapping any two direction parameter vectors $\pmb{\psi}=(\phi, \theta)$ and $\pmb{\psi}^{\prime}=(\phi^{\prime}, \theta^{\prime})$ to a real number, such that the Gram matrix $\pmb{K}=\left[k(\pmb{\psi}_i, \pmb{\psi}_j)\right]$ is positive semi-definite (PSD) for any finite collection $ \left\{\pmb{\psi}_1, \dots, \pmb{\psi}_n \right\}$ of direction parameter vectors~\cite{Rasmussen2004}. Different choices for the kernel $k$ give rise to different GPs. 
The Rational Quadratic (RQ) kernel, known for its capacity to regulate the extent of multi-scale behavior, stands out as a more adaptive option (see Sec.~\ref{sec:res_abl}) when compared to the Radial Basis Function (RBF) kernel \cite{shawe2004kernel, williams2006gaussian} : 


\begin{equation}
k_\mathrm{RQ}(\pmb{\psi}_i, \pmb{\psi}_j) = \left(1+\frac{d_{ij}^2}{2\alpha l^2}\right)^{-\alpha}. 
\label{eq:rational_quadratic}
\end{equation}
Here, $l > 0$, $\alpha > 0$ are hyperparameters and $d_{ij}$ denotes the Euclidean distance, $\Vert\pmb{\psi}_i-\pmb{\psi}_j\Vert$, between the orientation parameters. Perhaps a more suitable choice for $d_{ij}$ would be the geodesic distance between the bearing vectors $\pmb{u}(\pmb{\psi}_i)$, $\pmb{u}(\pmb{\psi}_j)$, but that would lead to a non-PSD kernel function as shown in \cite{kernels}, to which a  close valid alternative  would be the Euclidean distance, $\Vert \pmb{u}(\pmb{\psi}_i)- \pmb{u}(\pmb{\psi}_j)\Vert$. 


For a sufficiently sized training set of $n$ 3D points on the surface of the object, we may train the GP model to explicitly predict surface shape in terms of distance for given query directions from the point $\pmb{C}$. 
Thus, via the GP prior, for a query direction parameter vector $\pmb{x}\in\mathbb{R}^2$, we can obtain a conditional distribution over the distance to the surface along the given direction, and a collection of surface (training) data, $(\pmb{\Psi}, \pmb{d})=\left( \left[\pmb{\psi}_1,\dots, \pmb{\psi}_M\right], \left[d_1, \dots, d_M\right]\right)$:
\begin{equation}
    q\left(d\vert\ \pmb{x}, \pmb{C};(\pmb{\Psi}, \pmb{d})\right) \sim \mathcal{N}\left(\mu_{d\vert\pmb{x}}, \sigma^2_{d\vert\pmb{x}}\right),
    \label{eq:distance_likelihood}
\end{equation}
where, 
\begin{equation}
\begin{split}
\mu_{d\vert\pmb{x}}&=\pmb{K}(\pmb{x}, \pmb{\Psi})\pmb{K}(\pmb{\Psi}, \pmb{\Psi})^{-1}\pmb{d}, \\ 
    \sigma^2_{d\vert\pmb{x}}&=k(\pmb{x}, \pmb{x})-\pmb{K}(\pmb{x}, \pmb{\Psi})\pmb{K}(\pmb{\Psi}, \pmb{\Psi})^{-1}\pmb{K}(\pmb{\Psi}, \pmb{x}).
\end{split}
\label{eq:likelihood_mean_and_covariance}
\end{equation}
To discharge notation, we will henceforth omit the training data from the likelihood expression, $q\left(d\vert\ \pmb{x}, \pmb{C}; (\pmb{\Psi}, d)\right)$, and simply use,  $q\left(d\vert\ \pmb{x}, \pmb{C}\right)$.

\begin{figure}[b]
  \centering

  \begin{subfigure}{0.20\columnwidth}
    \includegraphics[width=\linewidth]{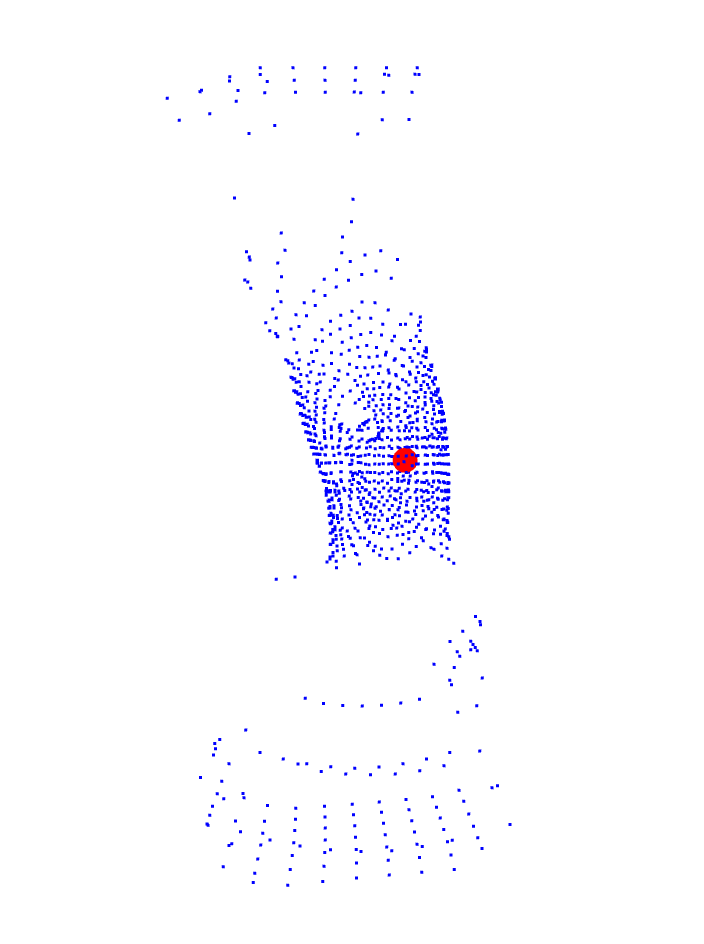}

  \end{subfigure}
  \hfill
  \begin{subfigure}{0.20\columnwidth}
    \includegraphics[width=\linewidth]{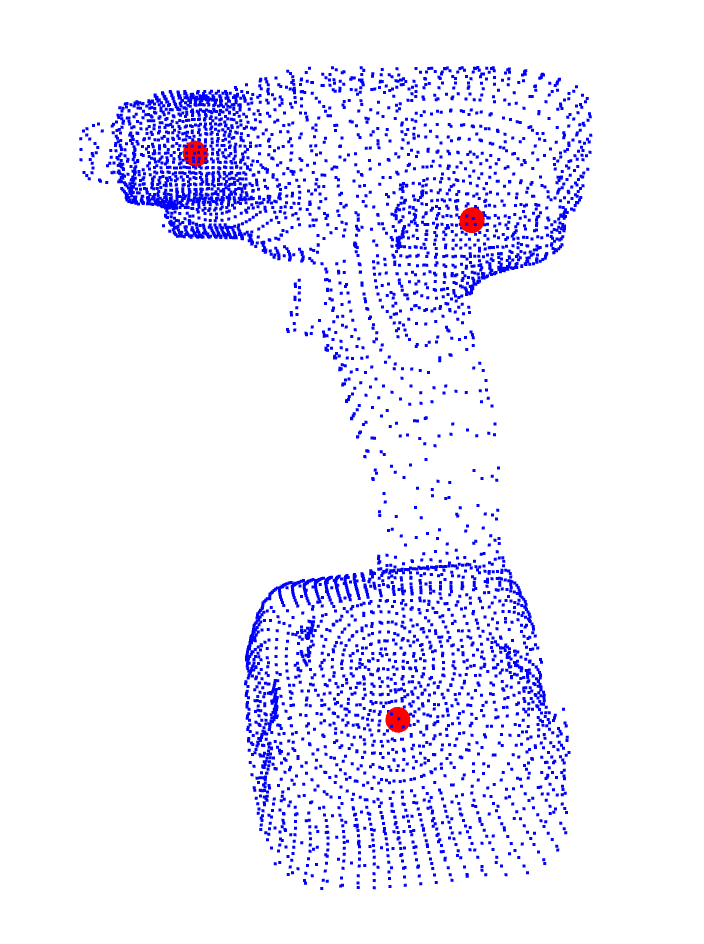}

  \end{subfigure}
  \hfill
  \begin{subfigure}{0.20\columnwidth}
    \includegraphics[width=\linewidth]{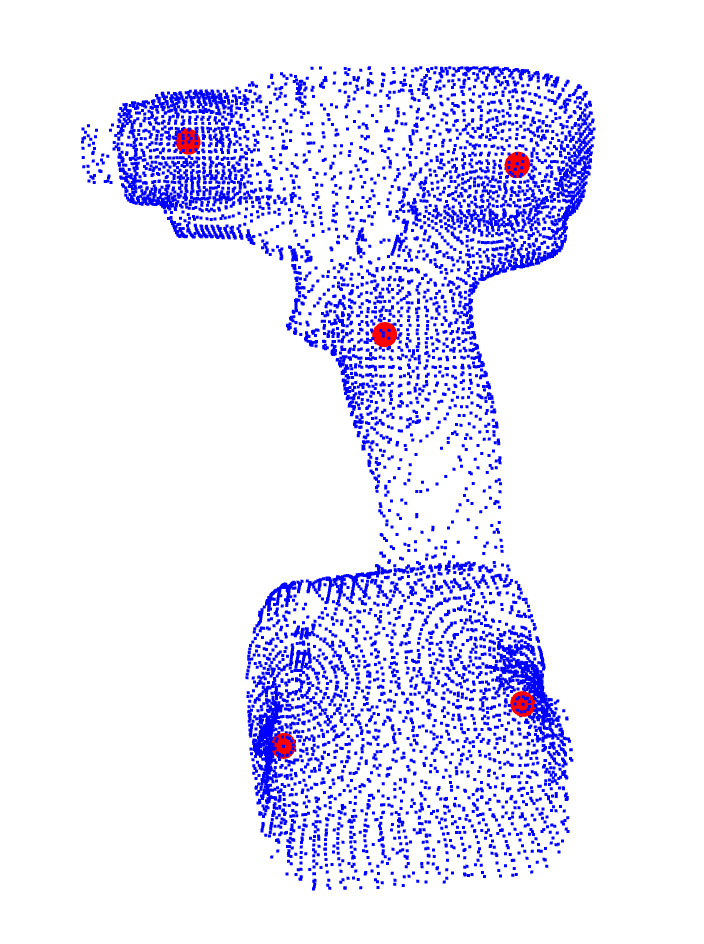}
    
  \end{subfigure}
  \hfill
  \begin{subfigure}{0.20\columnwidth}
    \includegraphics[width=\linewidth]{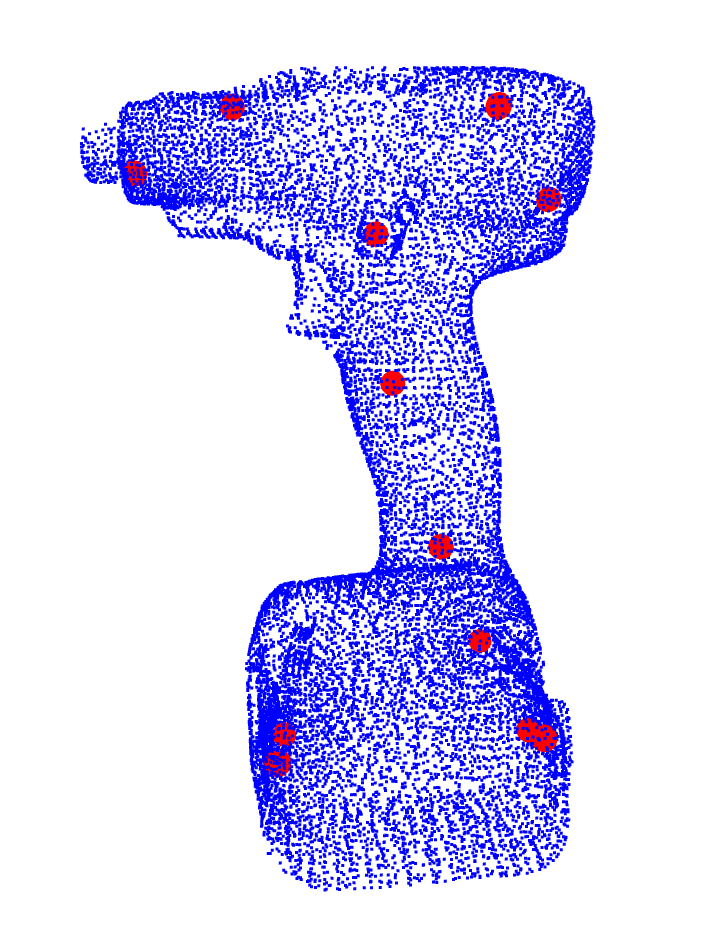}
    
  \end{subfigure}

  \caption{Ray-casting to the objects's surface (blue) from reference points (red) by retaining  the first intersection. More dense coverage achieved with increasing number of reference points (left-to-right).}
  \label{fig:centers}
\end{figure}

\subsubsection{Surface coverage with mixtures of likelihoods}

It becomes evident that for objects with more complex shapes, a single GP model will not suffice, because there would exist rays emanating from the reference point that intercept the exterior surface of the object in multiple points (e.g. Fig. \ref{fig:centers}). To cope with this issue, we introduce multiple reference points which we distribute preferably in the interior of the object in a way that achieves coverage of the entire surface. Interior reference points ensure that every direction corresponds to a valid training sample. 
A good strategy of obtaining such reference points would be distance-based clustering such as k-means 
 ~\cite{lloyd1982least} 
or expectation maximization (EM)~\cite{do2008expectation}, as it will induce a surface partition that will comprise loosely convex patches that are fully reachable from the respective cluster centers via ray-casting. 
For each of these reference points, a new GP prior is introduced. Therefore, the complete object surface is modeled by training multiple predictors of local surface patches with reference points at the cluster centers. 

\begin{figure}
  \centering
    \begin{subfigure}{0.24\columnwidth}
         \includegraphics[width=\textwidth]{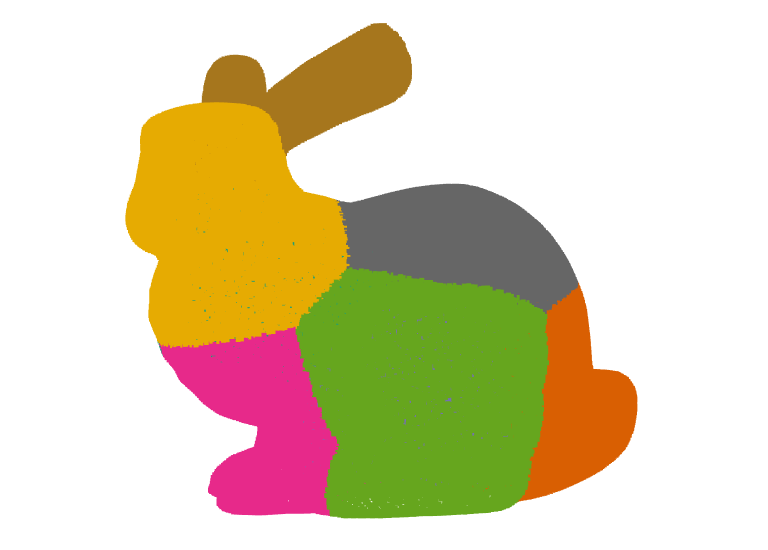}  
         \centering(a)
    \end{subfigure}
    \begin{subfigure}{0.24\columnwidth}
         \includegraphics[width=\textwidth]{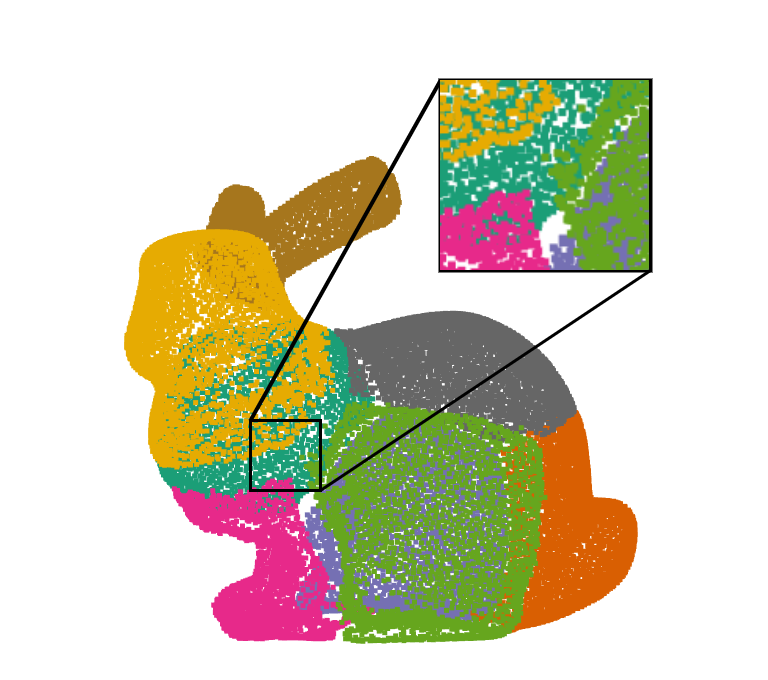} 
         \centering(b)
    \end{subfigure}
    \begin{subfigure}{0.24\columnwidth}
         \includegraphics[width=\textwidth]{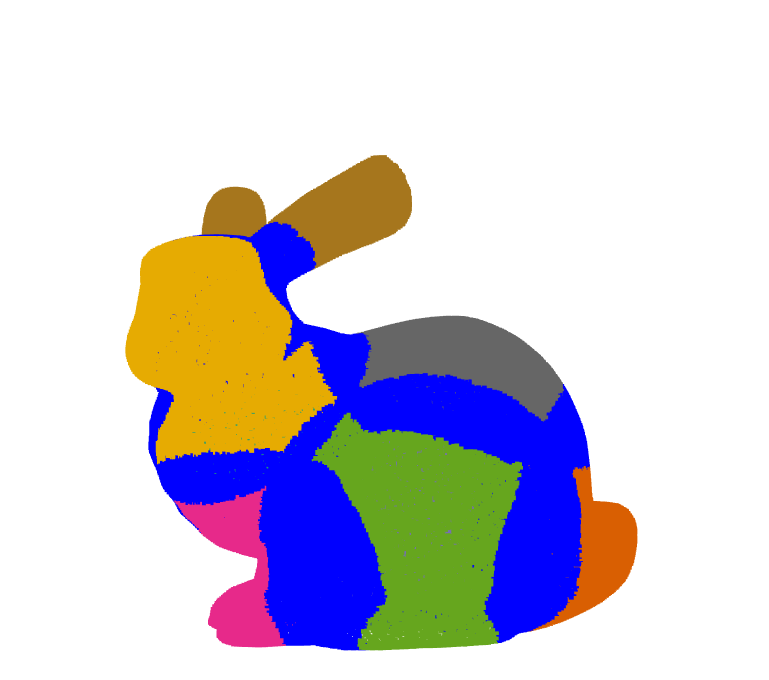}  
         \centering(c)
    \end{subfigure}
    \begin{subfigure}{0.24\columnwidth}
         \includegraphics[width=\textwidth]{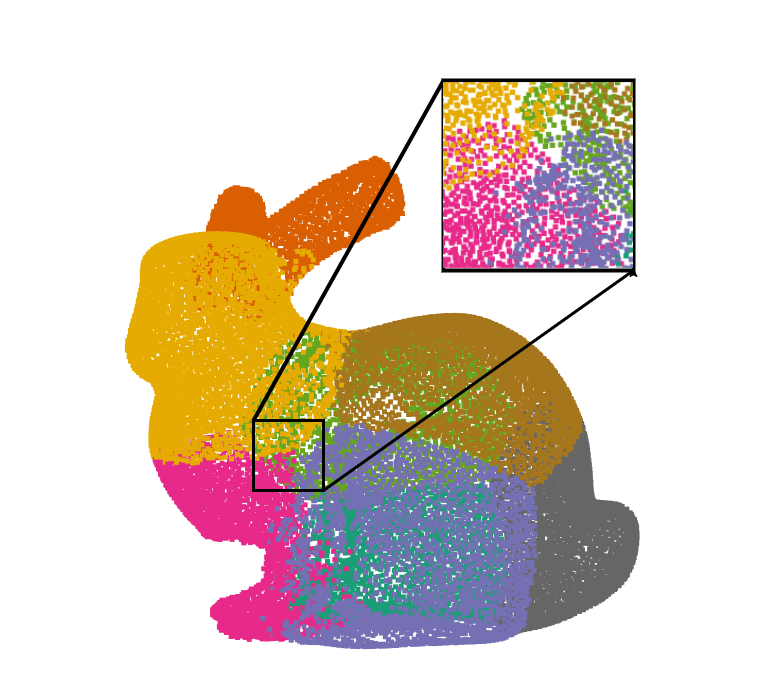} 
         \centering(d)
    \end{subfigure}

  \caption{Assignment of training points to clusters. (a)-(b) Ground truth - reconstructed without overlap. (c)-(d) Ground truth - reconstructed with overlap (blue).}

  \label{fig:bunny_inter_cluster}
\end{figure}

Formally, the idea is to model the likelihood of a 3D point $\pmb{P}$ belonging to the surface of the object $\pmb{o}$ as a mixture of normal distributions derived from the corresponding GP priors with reference points at the cluster centers $\pmb{C}_k$,
\begin{equation}
    p(\pmb{P}\vert \pmb{o})=\sum_{k=1}^K p\left(\pmb{P}\vert\pmb{C}_k; \pmb{o}\right)\pi(\pmb{C}_k; \pmb{o}),
    \label{eq:mixture_model}
\end{equation}
where we choose the likelihood $p\left(\pmb{P}\vert\pmb{C}_k; \pmb{o}\right)$ to be the GP prior based normal distribution of eq. \eqref{eq:distance_likelihood}, i.e.,
\begin{equation}
    p\left(\pmb{P}\vert\pmb{C}_k; \pmb{o}\right)=q\left(\left\Vert\pmb{P}-\pmb{C}_k\right\Vert \,\big\vert\, \pmb{x}_P, \pmb{C}_k\right),
    \label{eq:mixture_likelihood_from_gp}
\end{equation}
and $\pmb{x}_P$ is the direction parameter vector of $\pmb{P}$ with respect to the reference point $\pmb{C}_k$. Furthermore, the weighting probability $\pi(\pmb{C}_k; \pmb{o})$ is modeled as a softmax ratio based on the distance of $\pmb{P}$ from the reference point $\pmb{C}_k$:
\begin{equation}
    \pi(\pmb{C}_k; \pmb{o})=\frac{\exp\left(-\left(\pmb{P}-\pmb{C}_k\right)^T\pmb{Q}_k\left(\pmb{P}-\pmb{C}_k\right)\right)}{\sum_{l=1}^K \exp\left(-\left(\pmb{P}-\pmb{C}_l\right)^T\pmb{Q}_l\left(\pmb{P}-\pmb{C}_l\right)\right)},
    \label{eq:mixture_softmax_weights}
\end{equation}
where $\pmb{Q}_k$ is a covariance measure of the local space around the k-th cluster center (reference point) $\pmb{C}_k$ \footnote{Provided by the EM algorithm, whereas in the case of k-means we use the identity.}. In practice, this means that we reconstruct a query point $\pmb{P}$ by choosing the nearest center. This however can give rise to borderline errors in the assignment of query points to reference points. To cope with this problem, 
we impose inter-cluster overlap in the training (Fig. \ref{fig:bunny_inter_cluster}).

\label{ssect:surf_coverage_mixtures_distance_likelihoods}
\subsubsection{Distance likelihood and variance estimation}





Despite their efficiency, Gaussian processes are known to understimate variance, particularly away from data points~\cite{williams2006gaussian}. Since our confidence measure is an average conditional likelihood obtained from a Gaussian process, underestimated variance may skew the average. To alleviate this, we replace the estimated variance with one that we compute from the data using the GP predictions. We thus employ a modified likelihood distribution $\hat{q}$,
\begin{equation}
    \hat{q}\left(d\vert\ \pmb{x}, \pmb{C};(\pmb{\Psi}, \pmb{d})\right) \sim \mathcal{N}\left(\mu_{d\vert\pmb{x}}, \hat{\sigma}^2_{d\vert\pmb{x}}\right),
    \label{eq:distance_likelihood_hat}
\end{equation}
where, as in Eq. \eqref{eq:distance_likelihood}, the mean $\mu_{d\vert\pmb{x}}$ is the GP prediction, but the variance $\hat{\sigma}^2_{d\vert\pmb{x}}$, is the average squared deviation of the GP predictions from the ground truth points in a test set.

\label{ssect:dist_likelihood_test_data}
\subsection{Confidence scoring of pose estimates}

\label{ssect:confidence_scoring}

\subsubsection{Score calculation and pose consistency}
The concept behind our confidence measure essentially relies on using the 2D-3D correspondences derived from the pose estimation classifier and on the discrepancy of the back-projections of classified pixels (belonging to a known object) from the corresponding object template surface via the estimated pose. Consider a classifier, trained to recognize objects from a pool set $\mathcal{O}$, captured in images, and to map the corresponding pixels $\pmb{p}_i$, characterised as inliers (belonging to an object $o\in\mathcal{O}$) to 3D points $\pmb{P}_i$ in the object's local coordinate frame. Also, let $\pmb{T}\in\mathcal{SE}(3)$ be the pose prediction of the object by the classifier. Our confidence measure is the average weighted maximum probability of the back-projected points $\pmb{P}_i$ in Eq.~\eqref{eq:distance_likelihood_hat} amongst all possible reference points
\begin{equation}
    \mathcal{C}(\pmb{T};o) = \sum_{i=1}^N w_i \left(\underset{ k\in\left\{1,\dots,K\right\}}{\max}\hat{q}\left(\pmb{P}_i\vert \pmb{C}_k; o\right)\right),
    \label{eq:generic_confidence}
\end{equation}
where $w_i$ is a weight (typically, a probability) that encapsulates the classifier's self-assessed confidence about the quality of the pose estimate; clearly, if such a measure is not provided, then we choose $w_i=1/N$. 

\subsubsection{Bound on confidence score}
\label{sssect:confidence_from_residual_range}
Since our confidence measure is an average probability by definition, we may obtain a link between a desired margin, $\delta$, pertaining to the distance of a back-projected image points to the surface of the object. 
Thus, if, for each back-projected point $\pmb{P}_i$, the corresponding residual is bounded in absolute value by $\delta$, then we may obtain a lower bound for average confidence $\overline{\mathcal{C}}$ (see suppl. for derivations),
\begin{equation}
\overline{\mathcal{C}} \geq \frac{1}{\sqrt{2\pi}\delta^2}\sum_{i=1}^Nw_i\hat{\sigma}_{d\vert\pmb{x}}\left(1-\exp\left(-\frac{\delta^2}{2\hat{\sigma}_{d\vert\pmb{x}}^2}\right)\right),
\label{eq:confidence_bound}
\end{equation}
where $\hat{\sigma}_{d\vert\pmb{x}}^2$ is the estimated variance as per Sec. \ref{ssect:dist_likelihood_test_data}. With Eq.~\eqref{eq:confidence_bound} we are able to relate confidence values to a more tangible quantity such as distance.

%% file: sec/4_Experiments.tex
\section{Experiments}

\begin{table*}[!h]
\centering
\footnotesize 
\setlength{\tabcolsep}{4pt}
\begin{tabular}{l|cccc|cccc|cccc}
\toprule
& \multicolumn{4}{|c|}{\textit{Planes}} & \multicolumn{4}{|c|}{Chairs} & \multicolumn{4}{|c}{Sofas} \\
\cmidrule(lr){2-5} \cmidrule(lr){6-9} \cmidrule(lr){10-13}
& $d_C \downarrow$ & P $\uparrow$ & R $\uparrow$ & F $\uparrow$ & $d_C \downarrow$  & P $\uparrow$ & R $\uparrow$ & F $\uparrow$ & $d_C \downarrow$ & P $\uparrow$ & R $\uparrow$ & F $\uparrow$ \\
\hline
DeepSDF~\cite{park2019deepsdf}    & 0.79  & 90.2  & 85.5  & 87.6  & 0.81  & 84.2  & 88.2  & 86.0  & 0.57  & 89.0  & \textbf{90.0}  & 89.4  \\
DeepSDF (10k)     & 0.80    & \textbf{90.3}  & 84.5  & 87.2  & 0.90  &  84.5 & 88.2  & 86.1  & 0.56  & 65.0  & 60.8  & 62.5  \\
NKSR~\cite{huang2023nksr}     & 0.65  & 78.4  & 81.3  & 79.5  & 0.41  & 84.6  & \textbf{86.3}  & 85.0  & 0.45 & 79.6  & 82.5  & 80.8  \\
\hline
Ours        & \textbf{0.14}  & 87.4  & \textbf{93.6}  & \textbf{90.3}  & \textbf{0.21}  & \textbf{92.3}  & 86.5  & \textbf{89.2}  & \textbf{0.26}  & \textbf{92.4}  & 89.2  & \textbf{90.7}  \\
\bottomrule
\end{tabular}
\caption{Comparison with NKSR~\cite{huang2023nksr} and DeepSDF~\cite{park2019deepsdf} trained with 500K and 10K points per object across shapes of varying complexity for the planes, chairs, and sofas datasets from ShapeNetCore (120 models each). For the NKSR we used the pretrained model on ShapenetCore. We report $d_C$ (Chamfer distance), $P$ (Precision), $R$ (Recall), $F$ (F-score) computed using a sample of 30k points.}
\label{tab:shape_repr_metrics}
\end{table*}

\subsection{Experimental setup}
\PAR{Datasets.} We conduct our experiments on two datasets: ShapenetCore~\cite{shapenet2015} and IndustryShapes. ShapeNetCore is a densely annotated subset of the full ShapeNet corpus, containing 51,300 unique 3D models that span across 55 common object categories. We limit our experiments to 3 representative object categories, namely \textit{Planes}, \textit{Chairs} and \textit{Sofas} and use 120 models per category. IndustryShapes is a domain-specific internal dataset containing a limited set of tools and components that are used in an industrial setting of car-door assembly (cf. supplementary), wherein accurate pose estimation is a key enabler for robotic grasping. For our experiments, we use the \textit{Screwdrivers} object category.

\PAR{Evaluation protocol.} We perform two series of experiments. In the first, we demonstrate the accuracy of the GP mixture model as a shape template. To this end, we use the following metrics: Chamfer distance ($d_{C}$) to measure similarity to the ground-truth (GT) model, Precision (\textit{P}) to quantify the accuracy of the reconstructed model, Recall (\textit{R}) to quantify its completeness and F-score (\textit{F}) as an overall metric. Our second task is to evaluate the proposed confidence scoring of an estimated pose. In order to achieve this, we compute the reliable and widely used, average distance metric (ADD)~\cite{Hinterstoisser2012ACCV} and calculate the correlation between ADD and the proposed confidence values using Spearman's rank correlation coefficient~\cite{corder2014nonparametric}. A more detailed presentation of the metrics can be found in the supplementary material.

\PAR{Data preparation.} We prepare our input data following the methodology of DeepSDF~\cite{Park_2019_CVPR} to allow for a fair comparison. In particular, we first normalize the input model to the unit sphere. Subsequently, we sample a dense point cloud from the model's mesh surface by casting rays from a set of virtual cameras placed on a Fibonacci grid~\cite{swinbank2006fibonacci} onto the surface of the sphere. Our training and testing input is formed by sub-sampling two completely distinct sparse point clouds containing 10,000 and 30,000 points respectively.

\PAR{Optimization.} Our method uses GPyTorch~\cite{gardner2018gpytorch} for training and evaluating the GP models. We use the Adam optimizer~\cite{kinga2015method} with an initial learning rate of 0.1 and employ a learning rate scheduler to ensure stable convergence (see supplementary material for more implementation details).

 \subsection{Main results}
 \label{sec:eval_analysis}

 \begin{figure*}[htb]
    \centering
    \includegraphics[width=0.7\textwidth]{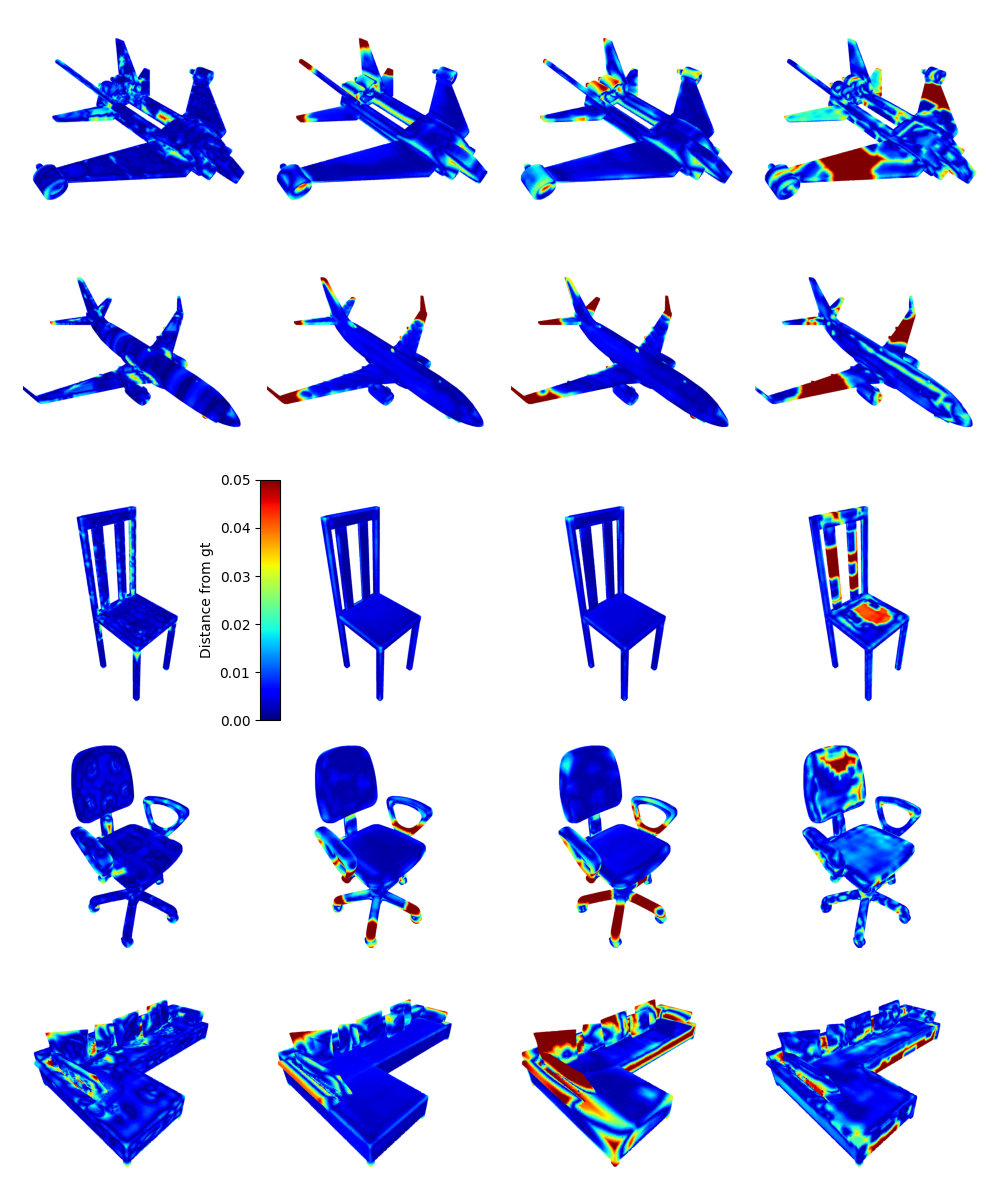}
    \caption{Qualitative comparison of shape representation on the ShapeNetCore dataset. The heatmap quantifies the distance from the ground truth point-cloud to the reconstructed; superimposed into the ground truth point cloud. We demonstrate two examples from the airplanes, chairs and sofas categories with each method. From left to right ours, DeepSDF~\cite{park2019deepsdf}, DeepSDF (10k), NKSR~\cite{huang2023nksr}}
    \label{fig:heatmap_results}
\end{figure*}

\PAR{Object representation.} Table \ref{tab:shape_repr_metrics} presents a quantitative comparison between our proposed shape representation and two state-of-the-art competitors: DeepSDF~\cite{park2019deepsdf} and NKSR~\cite{huang2023nksr}. Our method achieves the lowest Chamfer distance and over 90\% F-score across the board. Similarly, competitive Precision and Recall are attained for all object categories, with a remarkable 93.6\% Recall for the \textit{Planes} class. As the qualitative comparison of Fig. \ref{fig:heatmap_results} illustrates, our representation is able to capture the majority of the shape details in a robust manner, while competitors either over-smooth important shape features or miss them completely. It should be noted that we follow an intrinsically different approach. While DeepSDF and NKSR focus on generalization across different objects, the former by learning the latent space of shapes and the latter by combining the conditioning of kernel parameters on the data with kernel ridge regression, our method is more adept in representing details of specific objects.

\begin{figure}[!ht]
    \centering
    \begin{minipage}{0.30\columnwidth}
        \includegraphics[width=\textwidth]{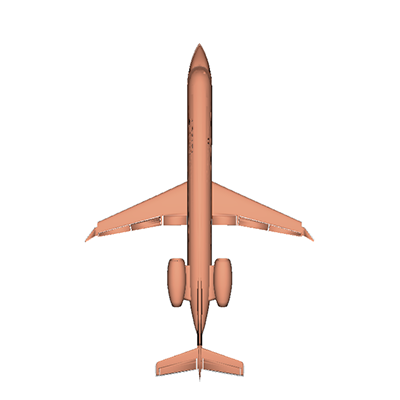}
        \subcaption*{ADD: 0.015, C: 0.78}
    \end{minipage}
    \begin{minipage}{0.30\columnwidth}
        \includegraphics[width=\textwidth]{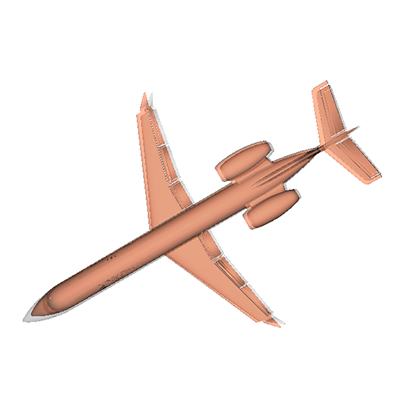}
        \subcaption*{ADD: 0.06, C: 0.46}
    \end{minipage}
    \begin{minipage}{0.30\columnwidth}
        \includegraphics[width=\textwidth]{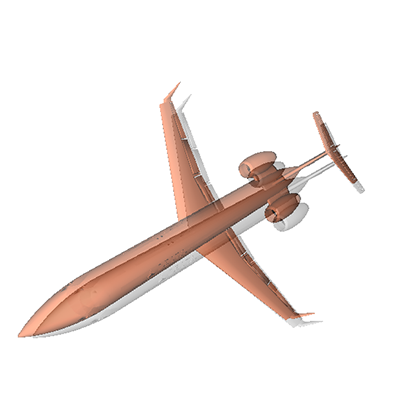}
        \subcaption*{ADD: 0.07, C: 0.16}
    \end{minipage}

    \begin{minipage}{0.30\columnwidth}
        \includegraphics[width=\textwidth]{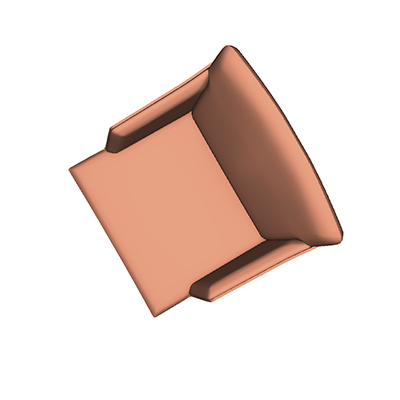}
        \subcaption*{ADD: 0.003, C: 0.79}
    \end{minipage}
    \begin{minipage}{0.30\columnwidth}
        \includegraphics[width=\textwidth]{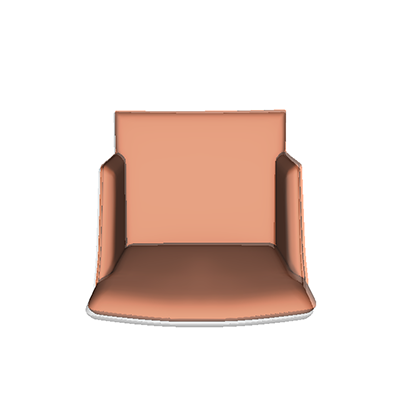}
        \subcaption*{ADD: 0.018, C: 0.57}
    \end{minipage}
    \begin{minipage}{0.30\columnwidth}
        \includegraphics[width=\textwidth]{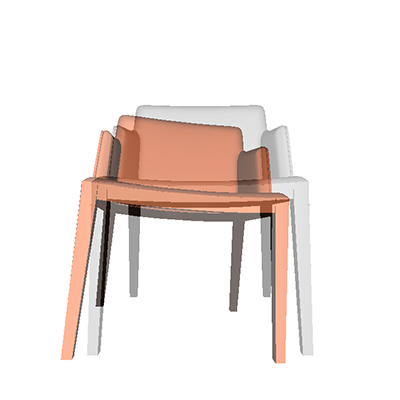}
        \subcaption*{ADD: 0.157, C: 0.10}
    \end{minipage}

    \begin{minipage}{0.30\columnwidth}
        \includegraphics[width=\textwidth]{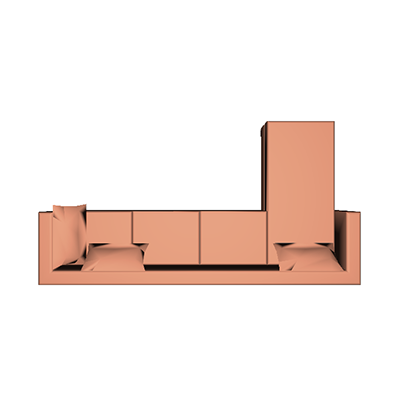}
        \subcaption*{ADD: 0.002, C: 0.80}
    \end{minipage}
    \begin{minipage}{0.30\columnwidth}
        \includegraphics[width=\textwidth]{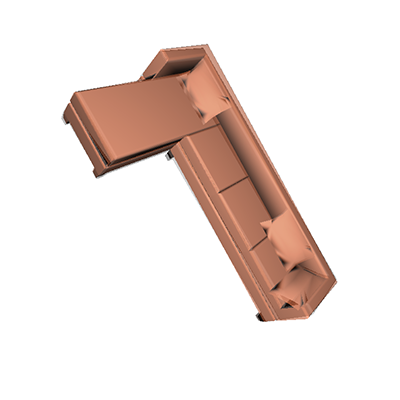}
        \subcaption*{ADD: 0.024, C: 0.55}
    \end{minipage}
    \begin{minipage}{0.30\columnwidth}
        \includegraphics[width=\textwidth]{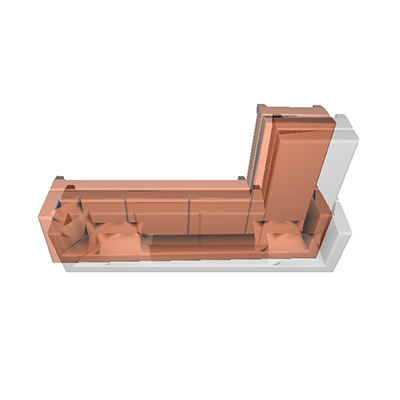}
        \subcaption*{ADD: 0.134, C: 0.08}

    \end{minipage}

        \begin{minipage}{0.30\columnwidth}
        \includegraphics[width=\textwidth]{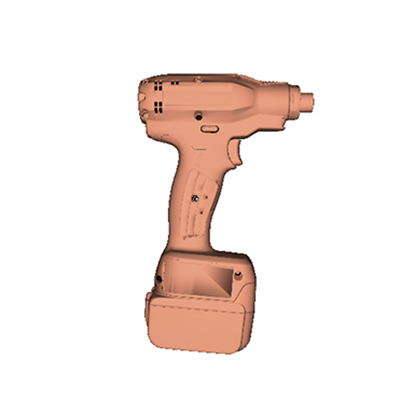}
        \subcaption*{ADD: 0.002, C: 0.81}
    \end{minipage}
    \begin{minipage}{0.30\columnwidth}
        \includegraphics[width=\textwidth]{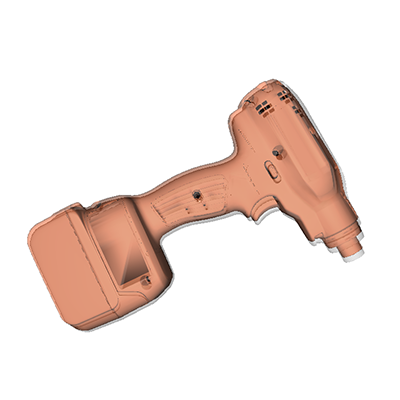}
        \subcaption*{ADD: 0.029, C: 0.59}
    \end{minipage}
    \begin{minipage}{0.30\columnwidth}
        \includegraphics[width=\textwidth]{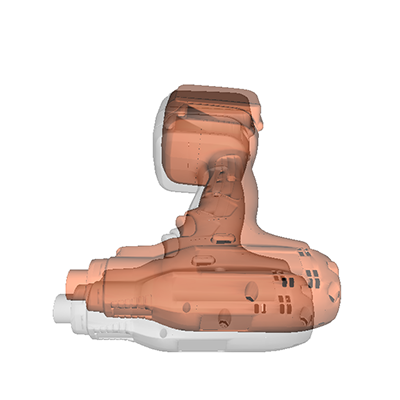}
        \subcaption*{ADD: 0.11, C: 0.08}
    \end{minipage}
    
    \caption{ Estimated poses (orange) of different 3D models from each object class derived synthetically with additive noise from rendered views and 3D-2D correspondences (cf. supplementary). The difference to ground truth is visualized. The ADD metric and the proposed confidence metric are computed for each case.}
    \label{fig:add_conf_poses}
\end{figure}

\begin{figure}[h]
\centering
\includegraphics[width=0.475\linewidth]{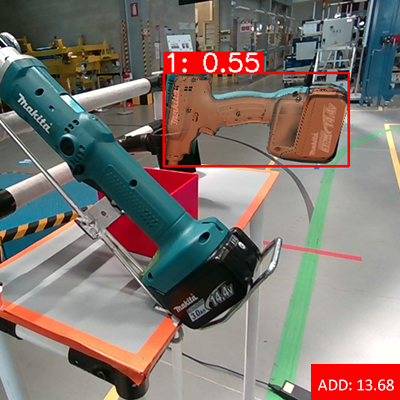}
\hspace{1pt}
\includegraphics[width=0.475\linewidth]{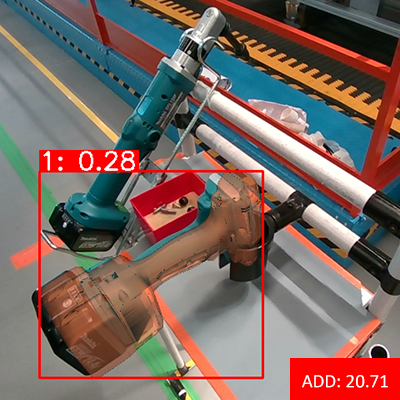}
\includegraphics[width=0.475\linewidth]{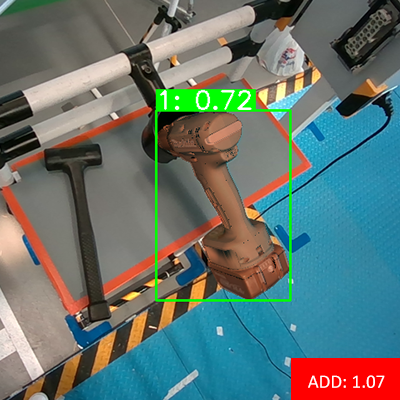}
\hspace{1pt}
\includegraphics[width=0.475\linewidth]{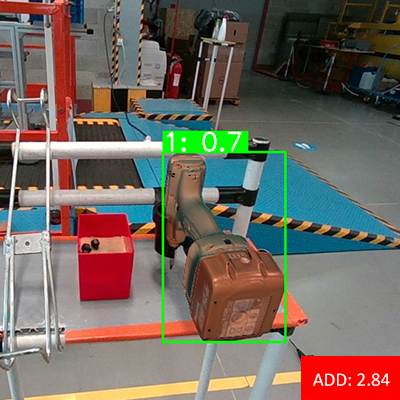}

\caption{Accepted (green) and rejected (red) pose estimates from industrial images based on our confidence metric with a threshold of 0.6 (via Eq. \eqref{eq:confidence_bound} for $\delta = 2mm$). The corresponding ADD can be seen in the lower right corner of each image.}
\label{fig:crf_poses}
\end{figure}

\PAR{Object pose evaluation and confidence scoring.} To assess the ability of the derived confidence score to accurately reflect the quality of the pose we rendered synthetic ground truth poses for all above object categories and projected a set of 3D points from each object on the image and thereafter, injected noise, $\epsilon$, into these image points according to a Gaussian mixture, $p(\epsilon) = p(\epsilon\vert o)\pi_o + p(\epsilon\vert \overline{o})(1-\pi_o)$ where $\pi_o$ is the outlier probability, $p(\epsilon\vert o)\sim\mathcal{N}(0, \sigma_o)$ and $p(\epsilon\vert \overline{o})\sim\mathcal{N}(0, \sigma_{\overline{o}})$ are the likelihood of noise for outliers and inliers, respectively, with $\sigma_{\overline{o}} \leq 2$ pixels and $\sigma_o > 10$ pixels. Subsequently, we computed the ADD and the confidence score for the estimated pose. Table~\ref{tab:mean_corr_coeff} presents a quantitative evaluation of the proposed confidence metric and relevant plots can be found in the supplementary. It is evident that our scoring mechanism exhibits a strong negative correlation with the ADD metric across all object categories, suggesting than an increase in ADD indicates a decrease in our confidence for the pose. Such a behavior is highly desirable for a scoring mechanism and demonstrates the capacity of our representation to provide a reliable evaluation metric for object pose estimates without directly comparing to ground truth data. In order to qualitatively illustrate the performance of our scoring approach, indicative results are shown in Fig.~\ref{fig:add_conf_poses} from all object categories for simulated poses with increasing noise, also including the ADD metric and the confidence score for each case. Additionally, we examine the pose estimates of 
\textit{Screwdriver} objects captured in real images inside an industrial environment. Fig.~\ref{fig:crf_poses} displays accepted (green) and rejected (red) poses based on our confidence metric. The superimposed object model further exemplifies the reliability of the pose scores in a challenging real-world setting.

\begin{table}[!h]
\centering
\footnotesize 
\setlength{\tabcolsep}{4pt}
\begin{tabular}{l|c|c|c|c}
\toprule & \textit{Planes} & \textit{Chairs} & \textit{Sofas} & \textit{Screwdrivers} \\
\hline
Mean Correlation Coefficient & -0.73 & -0.81 & -0.84 & -0.81 \\
\bottomrule
\end{tabular}
\caption{Correlation Analysis between ADD and the proposed confidence metric. The mean correlation coefficient for each object class is presented, indicating strong negative correlation.}
\label{tab:mean_corr_coeff}
\end{table}

\begin{figure}[!h]
\centering
  \includegraphics[width=0.9\columnwidth]{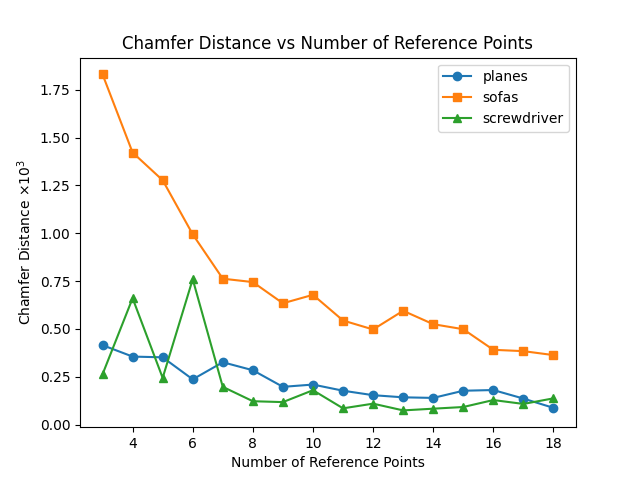}
  \caption{Ablation study for various number of reference points vs shape accuracy based on CD for the three object classes of the \emph{ShapeNetCore} dataset.}
  \label{fig:ablation}
\end{figure}

\subsection{Ablation studies}
\label{sec:res_abl}

\begin{figure}[h]
  \centering 
  \begin{subfigure}[t]{0.28\columnwidth}
      \includegraphics[width=\linewidth]{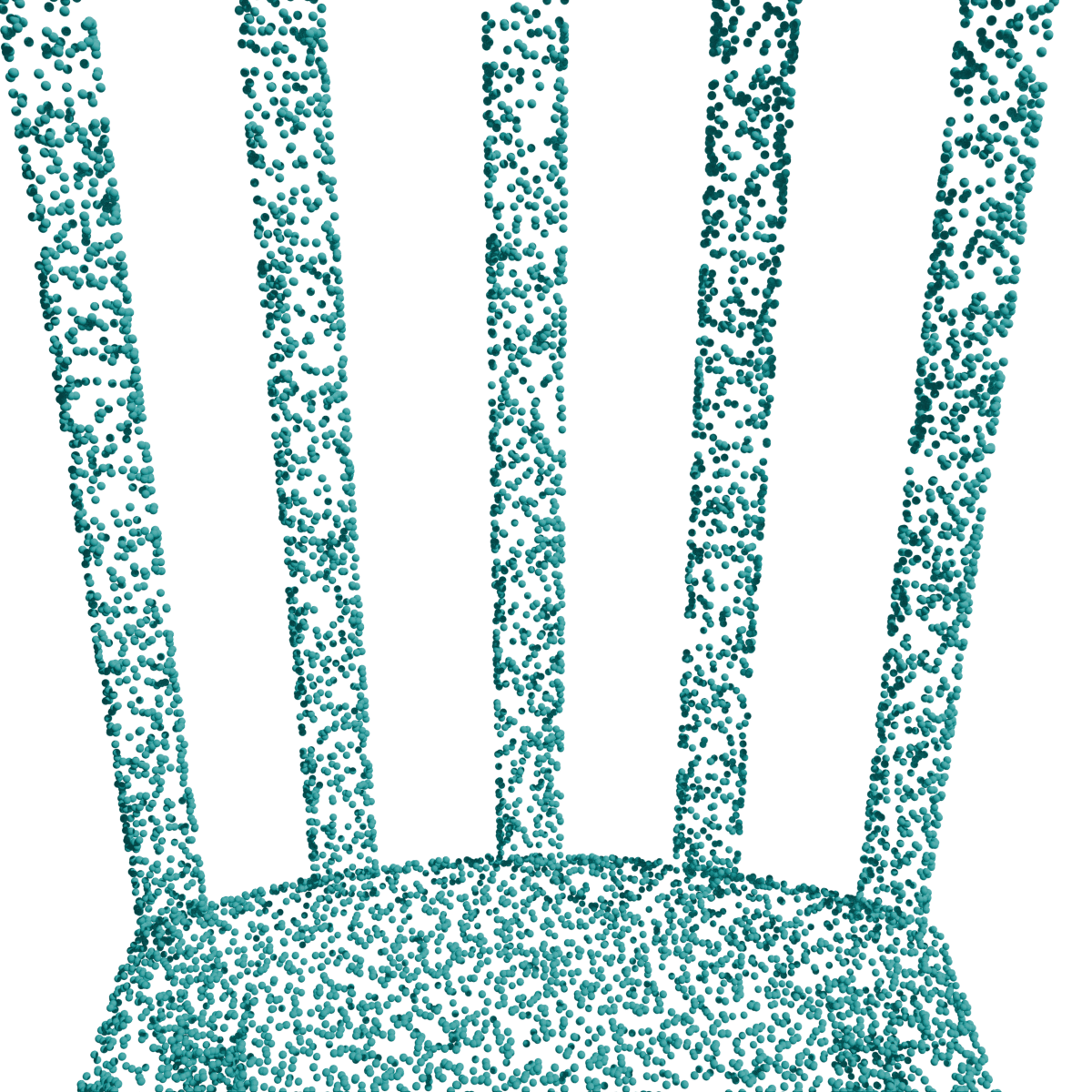}
    \caption{}
  \end{subfigure}
  \begin{subfigure}[t]{0.28\columnwidth}
    \includegraphics[width=\linewidth]{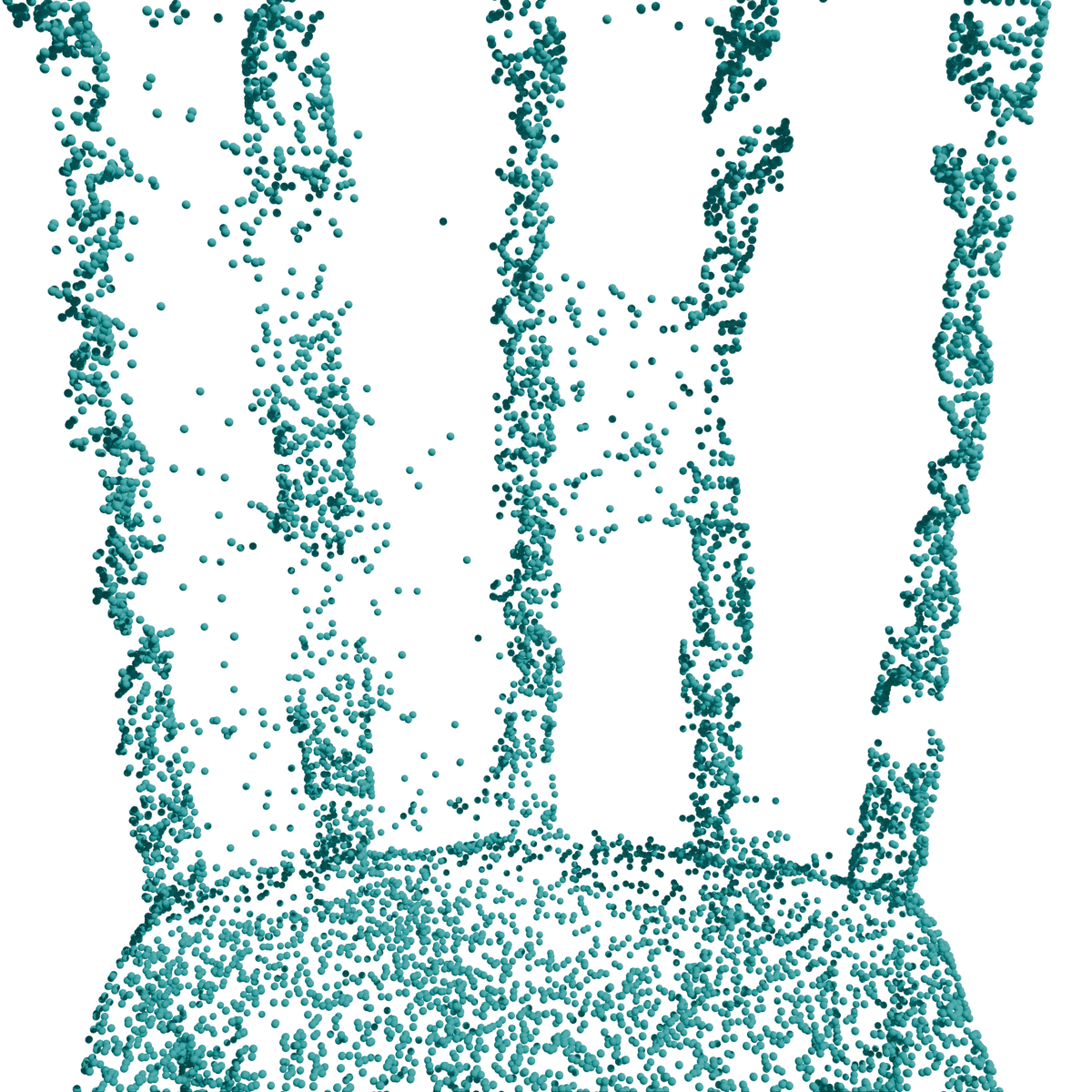}
    \caption{}
    \label{subfig:chair_e}
  \end{subfigure}
  \begin{subfigure}[t]{0.28\columnwidth}
    \includegraphics[width=\linewidth]{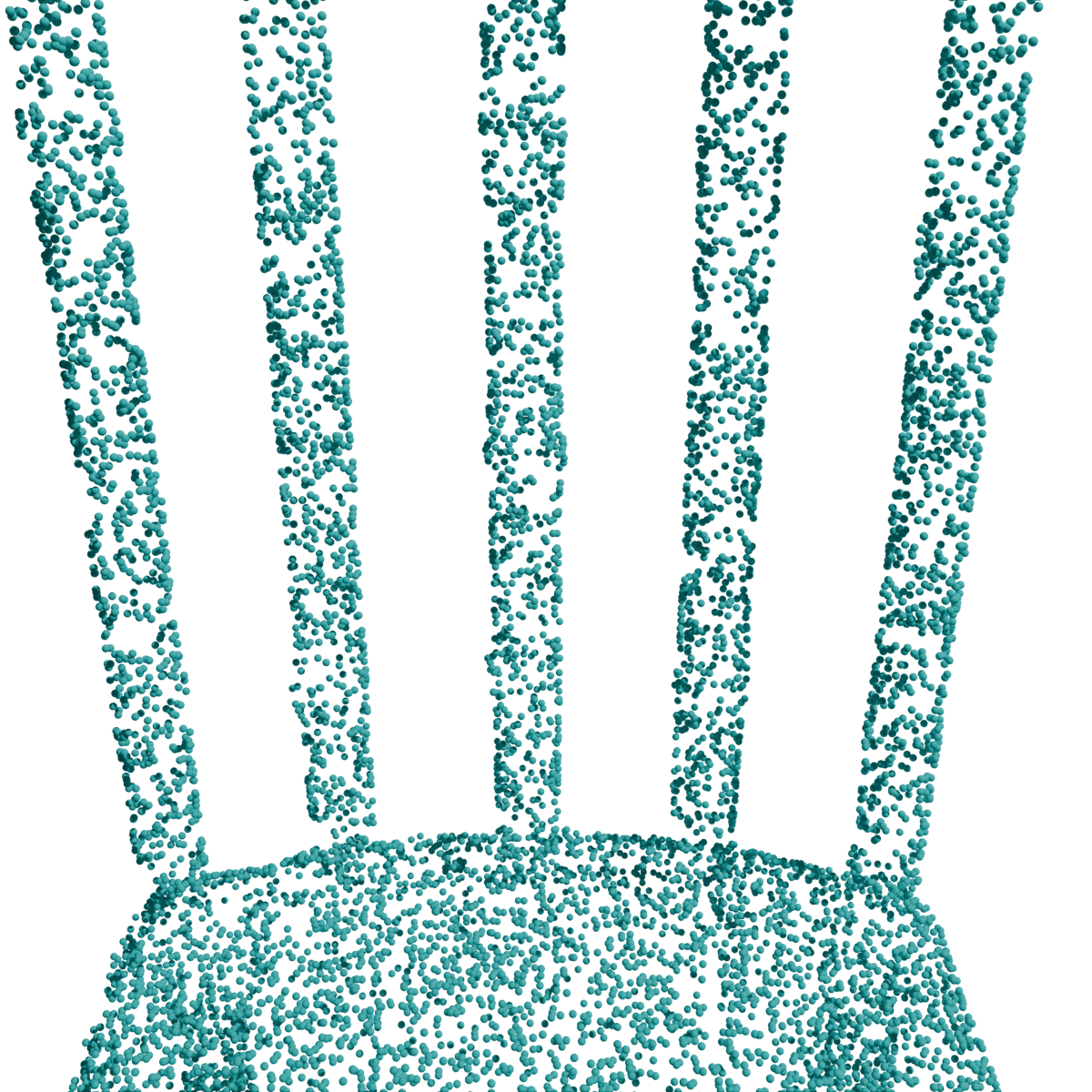}
    \caption{}
    \label{subfig:manual_chair_centers}
  \end{subfigure}

  \caption{(a) Ground truth, (b) Results obtained from automatically estimating reference point locations (16) with k-means, (c) Results based on manually defined reference points (32).}
  \label{fig:chair_manual_automatic}
\end{figure}

\PAR{Reference points.} We first study the impact that the multitude and the positioning of the reference points might have in the template's fidelity. Regarding the amount of reference points, Fig.~\ref{fig:ablation} illustrates that beyond a cutoff, the representation quality degrades significantly (i.e. Chamfer distance). As for their positioning, Fig.~\ref{fig:chair_manual_automatic} shows that the automatic reference points placement (using k-means) is inferior to a manual one for complex objects, but still adequate for our model to capture the object's topology (see also Sec.~\ref{sec:res_limits}). 

\PAR{Kernel choice.} We also experiment with a number of kernel types~\cite{Rasmussen2004} ranging from a third degree polynomial to periodic, linear, RBF, RQ and Matern. Our results (cf. supplementary) show that the RQ kernel exhibits superior performance for all object categories.

\subsection{Limitations}
\label{sec:res_limits}
Aside the number of points that impact the template fidelity (cf.~Sec.~\ref{sec:res_abl}), the locations of the reference points may also influence the quality of the representation. Overall, perturbing the number of reference points and their locations could lead to improvements, particularly when dealing with complex objects. For instance, in the case of the chair depicted in Fig.~\ref{fig:chair_manual_automatic}, comparing results obtained from clustering-based estimation of reference points (cf. Sec.~ \ref{ssect:lightweighht_shape_templates}) with those manually defined, a more accurate shape representation is derived for the latter. 

%% file: sec/5_Conclusion_Future_Work.tex
\section{Conclusion \& Future Work}
This paper puts forward a novel score for the quality of pose estimates of objects captured in images, based on geometric consistency with template shape. We also propose a method to build this template using Gaussian processes. Our results demonstrate that the representation is sufficiently accurate to facilitate the computation of a reliable confidence score that reflects the true quality of the estimated pose. In future work, we aim at endowing the shape representation algorithm with theoretical guarantees for full coverage of the object surface irrespective of shape complexity via learned mixtures models and fully automating the choice of reference points.

%% file: sec/Supplementary.tex
\section*{Supplementary}

\section{Lower bound for average confidence}

Consider the formula for confidence as the average likelihood of distance from reference points,
\begin{equation}
    \mathcal{C}(\pmb{T};o) = \sum_{i=1}^N w_i \left(\underset{ k\in\left\{1,\dots,K\right\}}{\max}\hat{q}\left(\pmb{P}_i\vert \pmb{C}_k; o\right)\right),
    \label{eq:generic_confidence_sup}
\end{equation}
For brevity, let $q^{\star}_i \sim \mathcal{N}(\mu_i^{\star}, {\sigma^{\star}}^2_i) \coloneq \underset{ k\in\left\{1,\dots,K\right\}}{\max}\hat{q}\left(\pmb{P}_i\vert \pmb{C}_k; o\right)$. Also, let $e_i = \Vert\pmb{P}_i-\pmb{C}_i^{\star}\Vert$ be the random variable of the difference between the GP prediction $\mu_i^{\star}$ and the actual distance of the query point from the reference point $\pmb{C}_i^{\star}$ along the direction of the query point, $\pmb{P}_i$. Then, the density of $e_i$ will be $q^{\star}$ translated at the origin. 
We now wish to find the average confidence value for $\vert e_i\vert<\delta$. We therefore integrate eq. \eqref{eq:generic_confidence_sup} from $-\delta$ to $\delta$ over $e_i$, i.e.,
\begin{equation}
    \begin{split}
        \overline{\mathcal{C}} = \frac{1}{2\delta}\int_{-\delta}^{\delta} \sum_{i=1}^N w_i \,q^{\star}_i\left(\pmb{P}_i\vert \pmb{C}_i^{\star}; o\right)\; de_i \\ =\frac{1}{2\delta} \sum_{i=1}^N w_i \int_{-\delta}^{\delta}q^{\star}_i\left(\pmb{P}_i\vert \pmb{C}_i^{\star}; o\right)\; de_i.
        \label{eq:confidence_integral}
    \end{split}
\end{equation}
We now consider the expression for the density $q^{\star}_i$,
\begin{equation}
q^{\star}_i\left(\pmb{P}_i\vert \pmb{C}_i^{\star}; o\right) = \frac{1}{\sqrt{2\pi}\sigma^{\star}_i}\exp\left(-\frac{e_i^2}{2{\sigma^{\star}}^2_i}\right).
\label{eq:density_exponential}
\end{equation}
Since we require that $\vert e_i \vert< \delta$, it follows that,
\begin{equation}
    1 > \text{sgn}(e_i)\cdot \frac{e_i}{\delta}.
\label{eq:inequlities}
\end{equation}
Combining the inequalities in eq. \eqref{eq:inequlities} with eq. \eqref{eq:density_exponential}, we get,
\begin{equation}
    \exp\left(-\frac{e_i^2}{2{\sigma^{\star}}^2_i}\right) \geq \text{sgn}(e_i)\cdot \frac{e_i}{\delta}\exp\left(-\frac{e_i^2}{2{\sigma^{\star}}^2_i}\right).
\label{eq:exponential_inequalities}
\end{equation}
Using eq. \eqref{eq:exponential_inequalities} with eq. \eqref{eq:confidence_integral}, we obtain a lower bound of the integral,
\begin{equation}
\begin{aligned}
    & \overline{\mathcal{C}} \geq  \frac{1}{2\sqrt{2\pi}\delta} \sum_{i=1}^N w_i \sigma^{\star}_i \left( \int_{-\delta}^{0} \frac{e_i}{{\sigma^{\star}}^2_i} \exp\left( -\frac{e_i^2}{2{\sigma^{\star}}^2_i} \right) \, de_i \right. \\
    & \left. \quad - \int_{0}^{\delta} \frac{e_i}{{\sigma^{\star}}^2_i} \exp\left( -\frac{e_i^2}{2{\sigma^{\star}}^2_i} \right) \, de_i \right) \\
    & \iff \overline{\mathcal{C}} \geq \frac{1}{\sqrt{2\pi}\delta} \sum_{i=1}^N w_i \sigma^{\star}_i \int_{\delta}^{0} \frac{e_i}{{\sigma^{\star}}^2_i} \exp\left( -\frac{e_i^2}{2{\sigma^{\star}}^2_i} \right) \, de_i.
\end{aligned}
\label{eq:confidence_integral_bound}
\end{equation}
Computing analytically the integral in eq. \eqref{eq:confidence_integral_bound} yields,
\begin{equation}
\overline{\mathcal{C}} \geq \frac{1}{\sqrt{2\pi}\delta^2}\sum_{i=1}^Nw_i
\sigma_i^{\star}\left(1-\exp\left(-\frac{\delta^2}{2{{\sigma}^{\star}}^2_i}\right)\right).
\label{eq:confidence_bound_sup}
\end{equation}

\section{Training Details}

\subsection{Hyperparameters}

To train our model to represent the shape of a specific object, we use $N$ sets of $S_i$ input samples, where $N$ denotes the number of reference points and $S_i$ represents the number of input samples assigned to reference point $i$. We treat $S_i$ as a single batch of points, which is then fed into the Gaussian Process (GP) model. Given that we use a sparse set of training points (e.g., 10,000 points in total), this typically poses no issues. Additionally, we set the default number of training iterations to 250.

\subsection{Loss function}

In a typical GP setup, the objective is to maximize the marginal likelihood, which represents the probability of the observed data under the GP prior and the likelihood. However, to ensure numerical stability, it is often preferable to maximize the Marginal Log Likelihood (MLL) or minimize the Negative Marginal Log Likelihood (NMLL). We employ NMLL as our loss function, which is defined as:
\begin{equation}
\begin{split}
    \mathcal{L} = - \log p(\mathbf{y} \mid \mathbf{X}) = \frac{1}{2} \mathbf{y}^\top (\mathbf{K} + \sigma^2 \mathbf{I})^{-1} \mathbf{y} + \\\frac{1}{2} \log |\mathbf{K} + \sigma^2 \mathbf{I}| + \frac{n}{2} \log 2\pi,
\end{split}
\end{equation}
where $y$ is the vector of observed data, $\mathbf{K}$ is the covariance matrix, $\sigma^2$ is the noise variance and, $\mathbf{I}$ is the identity matrix.

\subsection{Optimization}

We utilize the Adam optimizer with an initial learning rate of 0.1 and employ a learning rate scheduler to dynamically reduce the learning rate by a factor of 0.1 after 10 iterations without improvement in the loss function (i.e., patience of 10 iterations). Additionally, we experimented with second-order optimization algorithms such as L-BFGS with Strong Wolfe line search, but in practice, better results were obtained with the Adam optimizer.

\section{Hardware and software details}

Our pipeline is fully GPU-accelerated, leveraging GPytorch and CUDA. All experiments were conducted using a single RTX 4090 GPU with 24GB of VRAM. For an input sparse point cloud of 10,000 points, the GP models typically take about 30 seconds to train on the object, with slight variations depending on the number of reference points. Introducing overlapping regions increases the overall number of training samples.

\label{sec:hardware_specs}

\section{IndustryShapes Dataset}
\label{sec:industryshapes_dataset}
\begin{figure}[H]
    \centering
    \includegraphics[width=0.9\linewidth]{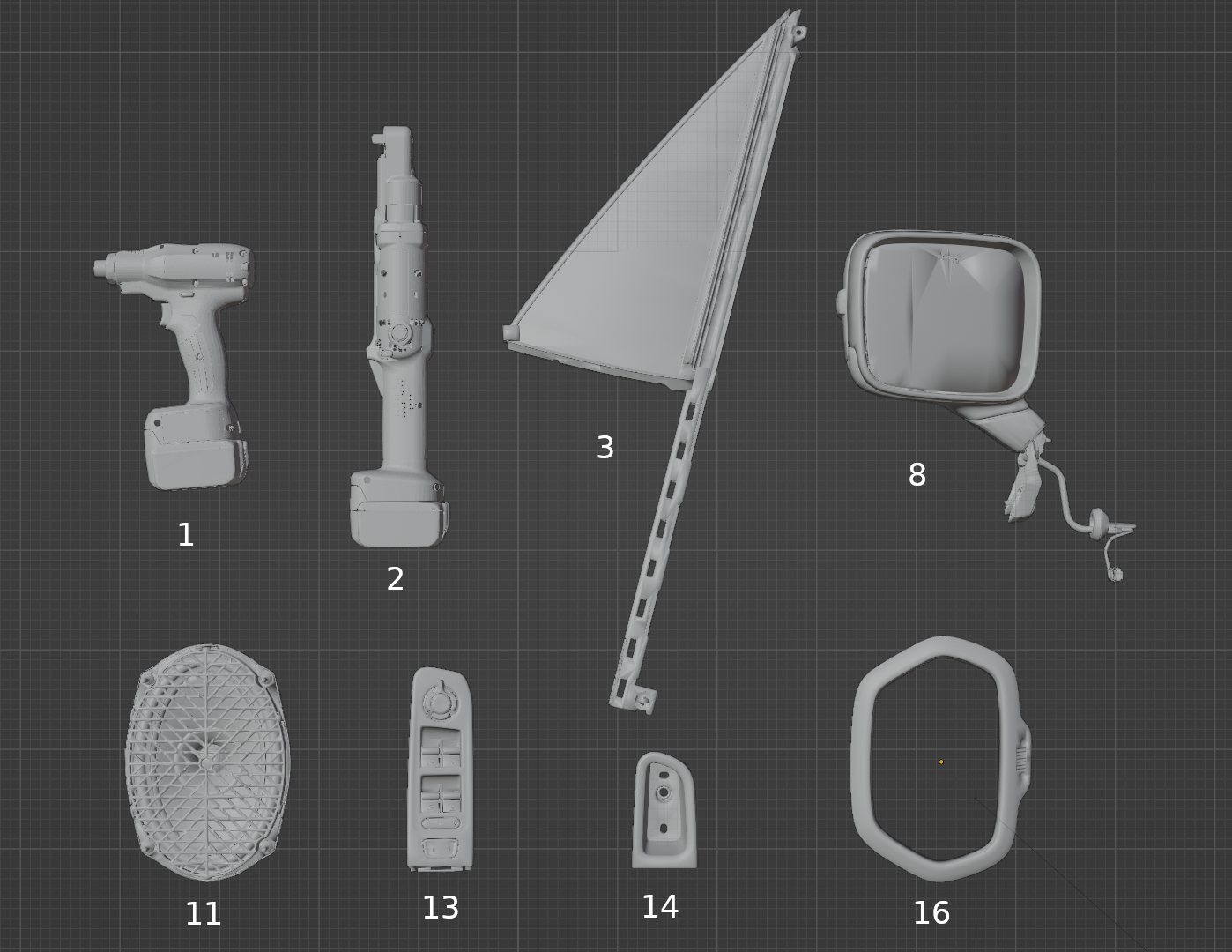}
    \caption{3D models of the IndustryShapes dataset}
    \label{fig:object_models}
\end{figure}

The IndustryShapes is a domain-specific dataset of tools and components that are used in an industrial setting of car-door assembly (Fig.~\ref{fig:object_models}). The objects feature challenging characteristics in terms of weak texture, surface properties and symmetries and are used in a human-robot collaboration scenario in which the robot has to detect, grasp and handover the objects to human assembly workers.

\section{Evaluation Details}

\subsection{Metrics}

To compute the evaluation metrics, we first sample points from both the ground truth mesh and the predicted mesh for methods that output meshes (e.g., NKSR, DeepSDF), as well as from the predicted point cloud generated by our method. Let $\mathbf{P_{gt}, P_{est}}$ 
denote the ground truth and estimated point sets, respectively. For the Chamfer distance computation,  we used a subset of 30,000 sampled points whereas for the rest of the metrics we used all the points of the test point-clouds (250k for the reported results). Specifically, for the NKSR method, we used the pre-trained model on ShapeNet and reconstructed the object’s surface using the same training point set used to condition our method.

\subsubsection{Chamfer Distance ($d_{C}$)} 
Chamfer distance is a commonly used metric in computer vision and machine learning, particularly for 3D reconstruction, to measure the similarity between two point sets. It can be defined in a one-way manner (i.e., how closely points from one set match those in the other set), which is non-symmetric and resembles precision or recall. However, it is most often evaluated in both directions. Given two point sets $\mathbf{P_{gt}, P_{est}}$ the chamfer distance is computed as:

\begin{equation}
\begin{split}
     \mathbf{d_C}(P_{gt}, P_{est}) = \frac{1}{|P_{gt}|} \sum_{x \in P_{gt}} \min_{y \in P_{est}} \|x - y\|^2 + \\ \frac{1}{|P_{est}|} \sum_{y \in P_{est}} \min_{x \in P_{gt}} \|x - y\|^2.
\end{split}
\end{equation}


\subsubsection{Precision ($P$)}

Precision measures the accuracy of the predicted points by quantifying how closely they match the ground truth. We compute the minimum distance between each predicted point $p_{est}$ and the ground truth set $P_{gt}$:
$$d_{p_{est} \rightarrow {P_{gt}}} = \min_{p_{gt} \in P_{gt}} \|p_{est} - p_{gt}\|.$$
By defining a distance threshold 
$\tau$, we can determine the percentage of predicted points whose distance to the ground truth falls within this threshold: 
\[
\mathbf{P}\left(\tau\right) = \frac{|\{p_{est} \in P_{est} :d_{p_{est} \rightarrow {P_{gt}}} < \tau\}| }{|P_{est}|}.
\]
In our experiments, we used a threshold of $\tau$ = 0.01 as all models are normalized to the unit sphere.

\subsubsection{Recall ($R$)}

Recall measures the completeness of the reconstruction, quantifying how well the ground truth points are represented in the predicted point cloud. Similar to Precision, we compute the minimum distance between each ground truth point $p_{gt}$ and the predicted set $P_{est}$:
$$d_{p_{gt} \rightarrow {P_{est}}} = \min_{p_{est} \in P_{est}} \|p_{gt} - p_{est}\|.$$
Using the same distance threshold $\tau$, Recall is the percentage of ground truth points that are within this threshold from the predicted points:
\[
\mathbf{R}\left(\tau\right) = \frac{|\{p_{gt} \in P_{gt} :d_{p_{gt} \rightarrow {P_{est}}} < \tau\}| }{|P_{gt}| }.
\]
We used the same distance threshold of $\tau$ = 0.01 for recall as well.

\subsubsection{F-score ($F$)}

The F-score combines Precision and Recall into a single metric, providing a more robust evaluation than either precision or recall alone. For instance, Recall can be maximized by overfilling the 3D space with points, while Precision can be artificially inflated by only retaining the most accurate predicted points. The F-score is defined as:

$$\mathbf{F} = \frac{2 \mathbf{P}(\tau) \mathbf{R}(\tau)}{\mathbf{P(\tau)} + \mathbf{R(\tau)}}.$$

\subsection{Analysis of confidence score}
\begin{figure}[h]
    \centering
    \includegraphics[width=0.9\columnwidth]{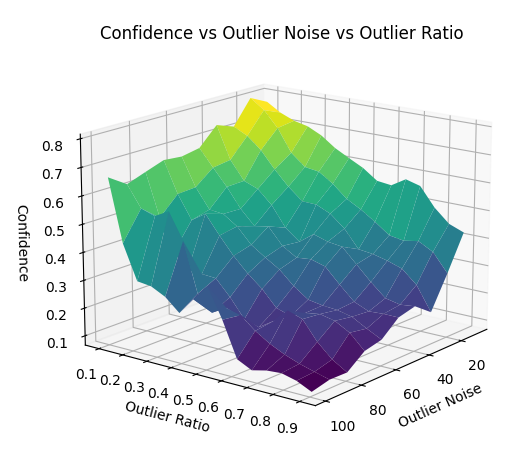}

    \caption{Confidence based on known poses with additional noise to simulate errors in object poses. Confidence vs outlier ratio and outlier noise.}
    \label{fig:conf_surf}
\end{figure}

\begin{figure}[h]
    \centering
    \includegraphics[width=0.9\columnwidth]{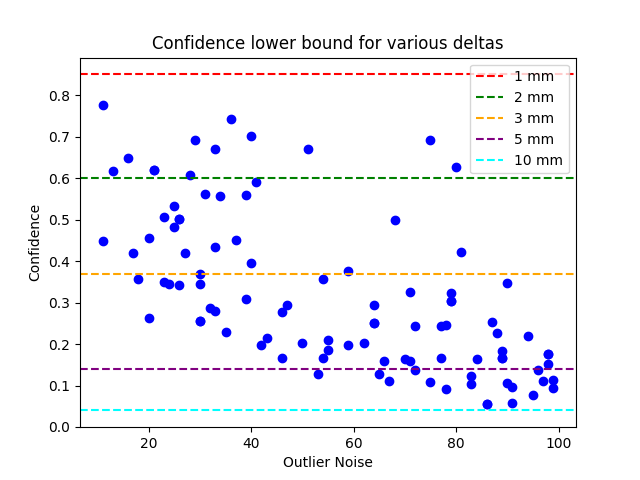}
    \label{subfig:deltas_conf_bound}
    
    \caption{Confidence based on known poses with additional noise to simulate errors in object poses. Confidence values for estimated poses (blue points) and confidence thresholds (horizontal lines) based on eq. \eqref{eq:confidence_bound_sup}, for various choices of $\delta$.}
    \label{fig:deltas_conf_bound}
\end{figure}

\begin{figure}[!h]
\centering
   \includegraphics[width=0.9\columnwidth]{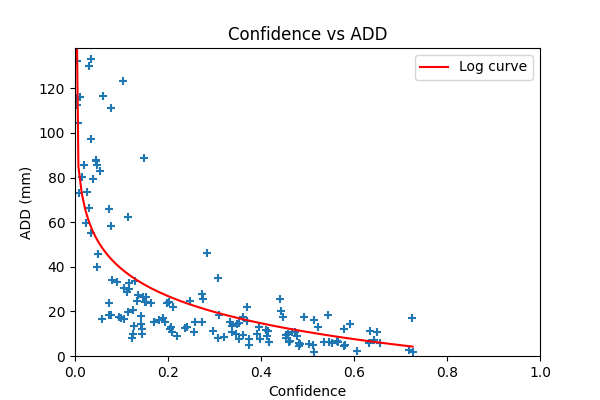}
   \captionof{figure}{Correlation analysis between our confidence score and ADD metric.}
   \label{fig:corr_conf_ADD}
\end{figure}

Further to the results reported in the main section of the paper (cf. sec. 4.2) we analysed the behavior of the confidence score for known ground truth poses simulating errors in object poses. Fig. \ref{fig:conf_surf} illustrates the relationship between confidence and noise, in terms of outlier ratio and outlier noise. It is evident that with increasing noise and outlier ratio the confidence score decrease. 
Fig. \ref{fig:deltas_conf_bound} illustrates how the choice of $\delta$ affects the confidence threshold and the acceptance of the estimated poses. Fig.~\ref{fig:corr_conf_ADD} illustrates the correlation analysis between our confidence score and the ADD metric for the case of the \emph{Screwdriver} object.

\subsection{Confidence metric evaluation through rendering of synthetic poses}

To generate the synthetic poses used for evaluating our confidence score, we first sample a fixed number of 3D points from the object’s surface (e.g., 500 points). Using predefined ranges of Euler angles  ($\phi$, $\theta$) and distance values, we iteratively compute a ground truth transformation matrix $R_{GT}$. These 3D points are then projected onto the image plane using fixed intrinsic parameters and $R_{GT}$ resulting in a set of 2D points, $P_{2D}$. Then, we introduce noise. Specifically, inlier noise is added with a fixed standard deviation ($\sigma_{\overline{o}} \leq 2$), while the probability of a point being an outlier is varied to simulate different levels of outlier contamination. Additionally, the random displacement of outliers ($\sigma_{o}$) is also varied. The resulting noisy 2D points, $\Tilde{P}_{2D}$, are used to estimate a noisy pose, $R_n$, by solving the Perspective-n-Point (PnP) problem with the original 3D points. Finally, we compute the ADD and the proposed confidence metrics. These metrics are derived using the noise pose $R_n$, with ADD specifically requiring the ground truth transformation matrix $R_{GT}$ for comparison.

\subsection{Kernel choice}

To identify a suitable kernel for our application that performs robustly across objects of varying classes and complexities, we conducted structured experiments testing a range of kernels, as detailed in Table \ref{tab:kernel_choice}. To ensure a fair comparison, all models were trained with eight fixed reference points per object, eliminating randomness in center initialization. For the Planes and Chairs categories, five models were trained, and the mean metrics for each category were evaluated. The results indicate that the RBF, Matérn, and Rational Quadratic (RQ) kernels perform well, with the RQ kernel achieving the best overall performance. The metrics reported include Chamfer distance, Precision, Recall, and F1-score.

\begin{figure}[!h]
\centering
   \includegraphics[width=0.9\columnwidth]{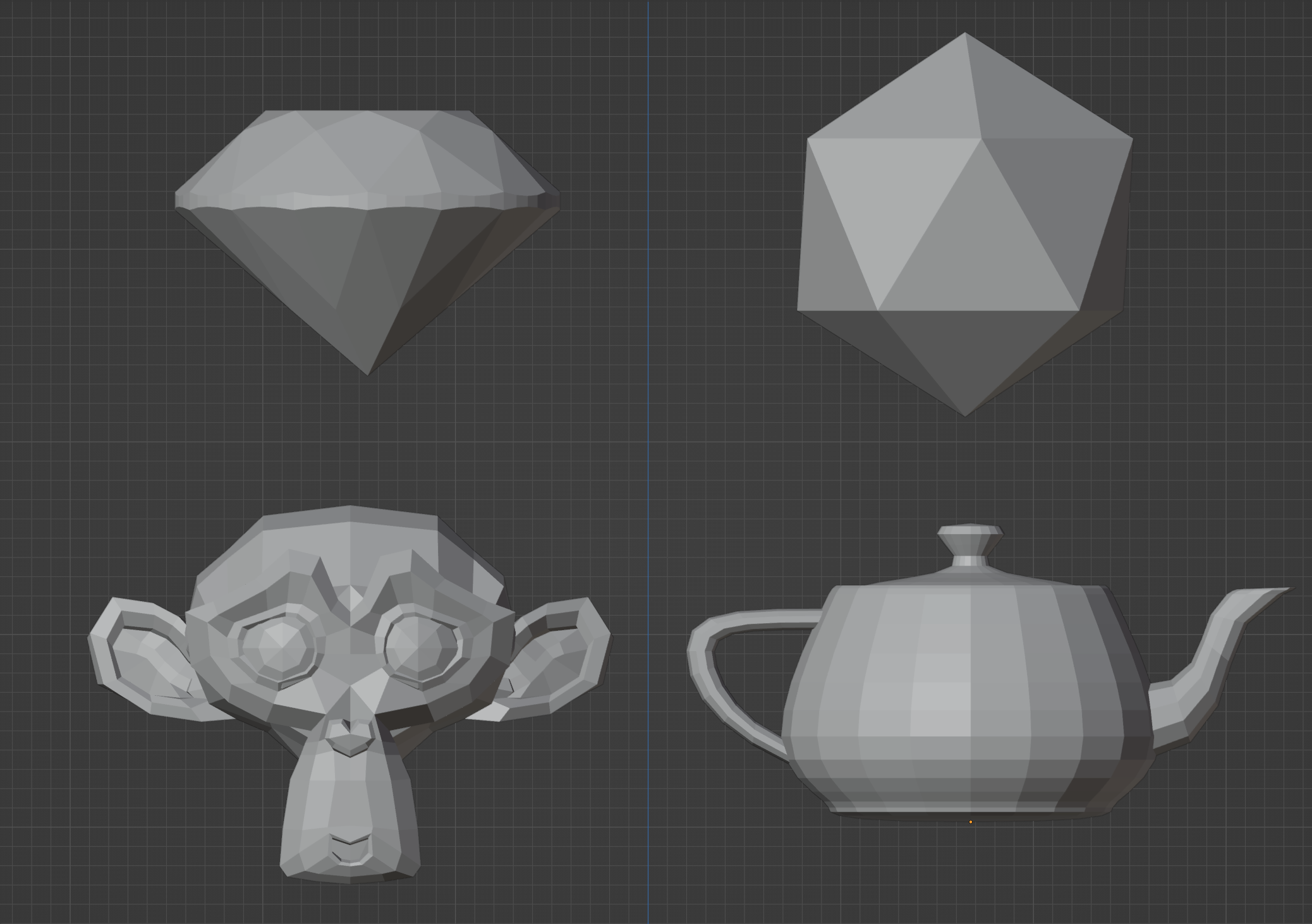}
   \captionof{figure}{Objects of various complexities used for the KDTree - COBRA GPs comparison. The objects of the first row are labeled as "low complexity" while the objects of the second row as "high complexity".}
   \label{fig:objects_kdtree}
\end{figure}

\subsection{Merits of Functional representation (COBRA) vs traditional storage (KDTree)}


We construct a functional representation of object shape using Gaussian Processes (GPs). It is a well known fact that functional representations tend to be more economic in terms of storage and lookup operations as opposed to traditional pointclouds. Furthermore, the key advantage of a functional representation lies in its ability to generalize from sparse input data points, approximating the object’s surface in a continuous manner. 

We showcase the advantages of our choice of functional representation (GPs), and evaluate performance on four objects: two with relatively simple geometry and two with more complex shapes. The objects, shown in Fig. \ref{fig:objects_kdtree}, were generated as base meshes using Blender software. These objects were selected for their controllable shape complexity, in contrast to the highly complex objects in the datasets used for our main results.

\begin{figure*}[!h]
\centering
   \includegraphics[width=0.7\textwidth]{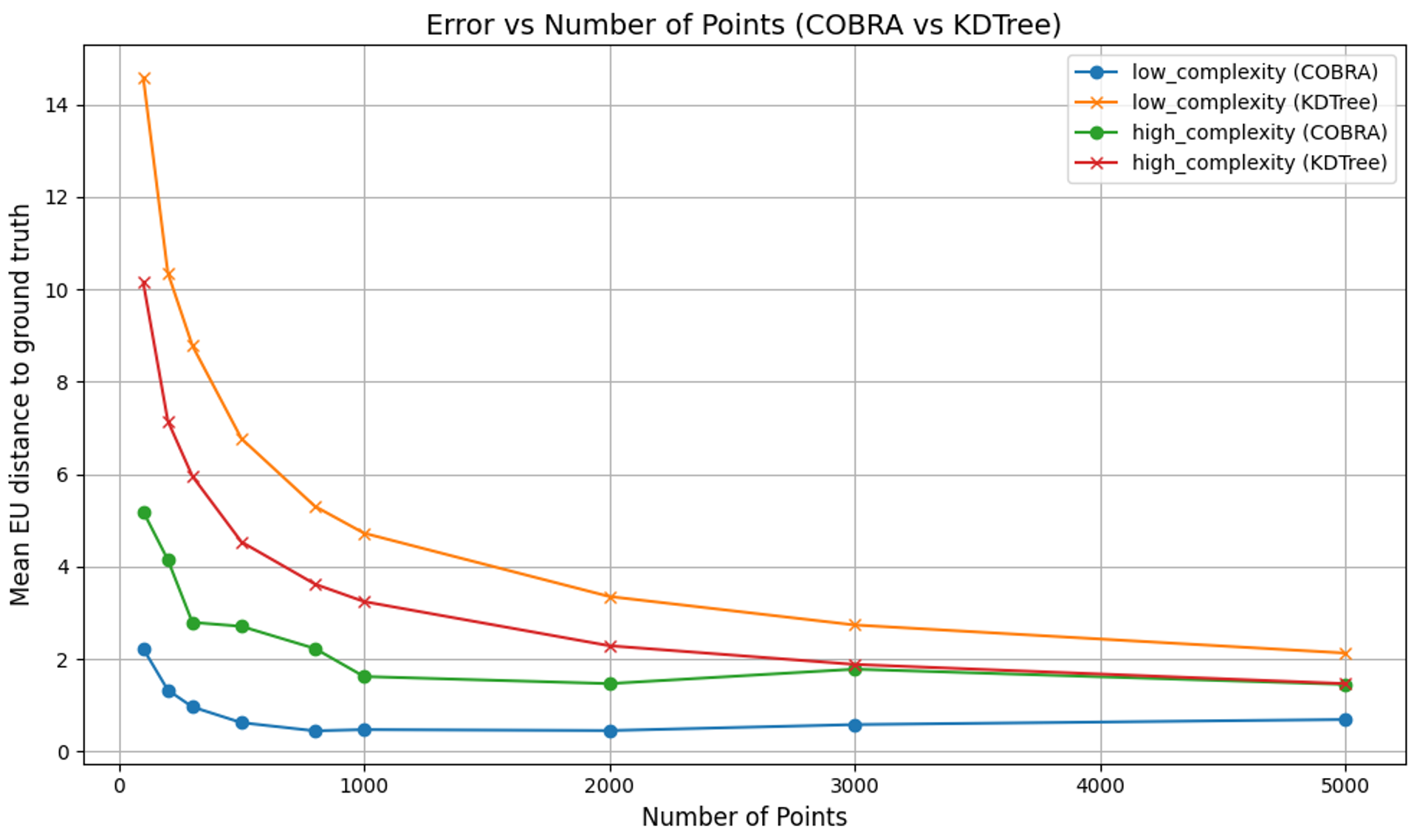}
   \captionof{figure}{COBRA vs. KDTree comparison in terms of accuracy achieved with sparse input point sets. Error ranges are normalized to objects scaled to fit within a sphere of radius 100.}
   \label{fig:KDTree_cobra}
\end{figure*}

For this study, we trained COBRA GPs for each object using varying numbers of input points, ranging from 100 to 5000. For comparison, we also created KD-Trees using the same input points and evaluated both methods on a test point cloud of 10,000 points. As illustrated in Fig. \ref{fig:KDTree_cobra}, GPs consistently outperformed KD-Trees, demonstrating better accuracy for both simple and complex models, particularly when input point clouds were sparse.

\begin{table*}[htbp]
    \centering
    \setlength{\tabcolsep}{3pt} 
    \renewcommand{\arraystretch}{1.2} 
    \begin{tabular}{lcccccccccccccccc}
        \toprule
        & \multicolumn{4}{c}{Planes} & \multicolumn{4}{c}{Chairs} & \multicolumn{4}{c}{Screwdriver} & \multicolumn{4}{c}{Bunny} \\
        \cmidrule(lr){2-5} \cmidrule(lr){6-9} \cmidrule(lr){10-13} \cmidrule(lr){14-17}
        & $d_C$ ↓ & P ↑ & R ↑ & F ↑ 
        & $d_C$ ↓ & P ↑ & R ↑ & F ↑
        & $d_C$ ↓ & P ↑ & R ↑ & F ↑ 
        & $d_C$ ↓ & P ↑ & R ↑ & F ↑ \\
        \midrule
        \textbf{Polynomial (p=3)} \newline & 1.56 & 35.3 & 58.0 & 43.7 
        & 3.35 & 26.7 & 40.4 & 31.9 
        & 1.69 & 32.6 & 32.8 & 32.7 
        & 2.71 & 23.3 & 31.8 & 26.9 \\
        \textbf{Periodic} 
        & 0.38 & 70.4 & 86.4 & 77.5 
        & 0.62 & 60.8 & 73.8 & 66.2 
        & 3.47 & 26.6 & 26.8 & 26.7 
        & 5.60 & 25.0 & 24.7 & 24.9 \\
        \textbf{Linear} 
        & 4.37 & 13.5 & 36.2 & 19.5 
        & 9.09 & 10.9 & 31.8 & 16.0 
        & 9.02 & 10.5 & 14.6 & 12.2 
        & 12.1 & 8.15 & 18.0 & 11.2 \\
        \textbf{RBF} 
        & 0.31 & 73.0 & 88.9 & 80.1 
        & 0.47 & 60.7 & 69.2 & 66.4 
        & 0.20 & 86.0 & 88.4 & 87.2 
        & 0.45 & 72.5 & 77.5 & 75.0 \\
        \textbf{RQ} 
        & \textbf{0.19} & \textbf{82.6} & \textbf{93.8} & \textbf{87.9} 
        & 0.51 & \textbf{68.4} & \textbf{82.9} & \textbf{74.5} 
        & \textbf{0.12} & \textbf{95.1} & \textbf{96.5} & \textbf{95.8} 
        & \textbf{0.33} & \textbf{81.8} & \textbf{87.8} & \textbf{84.7} \\
        \textbf{Matern} 
        & 0.26 & 76.3 & 91.8 & 83.3 
        & \textbf{0.43} & 64.7 & 79.2 & 70.8 
        & 0.15 & 92.0 & 93.6 & 92.8 
        & 0.39 & 77.5 & 82.8 & 80.1 \\
        \bottomrule
    \end{tabular}
    \caption{Ablation study of different kernels and their impact on reconstruction accuracy across various shapes. Chamfer distance is multiplied by $10^3$.}
    \label{tab:kernel_choice}
\end{table*}